\documentclass[11pt,letterpaper]{mystyle}
\usepackage{fancyhdr}
\usepackage{tcolorbox}

\usepackage[utf8]{inputenc} 
\usepackage[T1]{fontenc}
\usepackage[numbers]{natbib}
\usepackage{adjustbox}
\usepackage{breakurl}    
\usepackage{hyperref}
\usepackage[edges]{forest}
\usepackage{subcaption}
\usepackage{soul}
\usepackage{multirow}
\usepackage[utf8]{inputenc} 
\usepackage{booktabs}       
\usepackage{amsfonts}       
\usepackage{nicefrac}      
\usepackage{microtype}     
\usepackage{xcolor}        
\usepackage{colortbl}
\usepackage{enumitem}

\usepackage{xcolor}
\usepackage{amssymb}
\usepackage{wrapfig}
\usepackage{courier}
\usepackage{bxcoloremoji}
\usepackage{CJKutf8}

\usepackage{amsmath,amssymb,amsthm}

\newtheorem{definition}{Definition}
\newtheorem{lemma}{Lemma}
\newtheorem{proposition}{Proposition}

\definecolor{mypurple}{rgb}{0.878, 0.748, 0.996}
\definecolor{mydarkpurple}{rgb}{0.778, 0.648, 0.896}
\definecolor{myblue}{rgb}{0.830, 0.839, 0.993}
\definecolor{mygreen}{rgb}{0.821, 0.931, 0.862}

\fancypagestyle{headstyle}{
    \fancyhead[L]{
        \includegraphics[width=80pt]{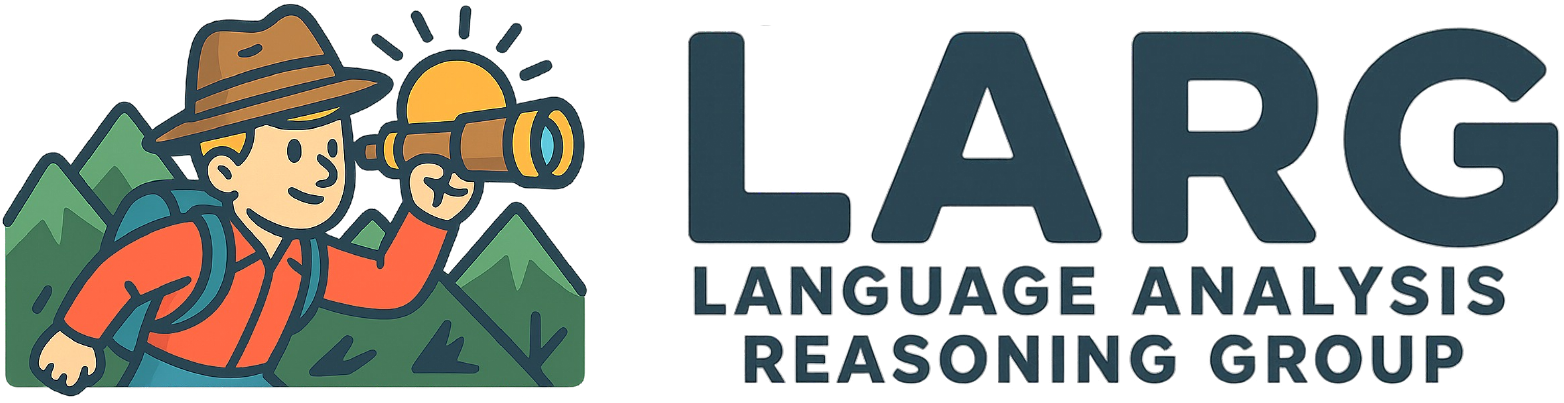}
    }
    \fancyhead[C]{}
    \fancyhead[R]{}

}

\definecolor{hidden-red}{RGB}{205, 44, 36}
\definecolor{hidden-blue}{RGB}{194,232,247}
\definecolor{hidden-orange}{RGB}{243,202,120}
\definecolor{hidden-green}{RGB}{34,139,34}
\definecolor{hidden-pink}{RGB}{255,245,247}
\definecolor{hidden-black}{RGB}{20,68,106}
\definecolor{purple}{RGB}{144,153,196}
\definecolor{yellow}{RGB}{255,228,123}
\definecolor{hidden-yellow}{RGB}{255,248,203}
\definecolor{tkcolor}{RGB}{224,223,255}
\definecolor{darkblue}{rgb}{0, 0.40, 0.75}

\hypersetup{colorlinks=true, citecolor=darkblue, linkcolor=darkblue, urlcolor=darkblue}
\tcbset{
  aibox/.style={
    width=\linewidth,
    top=8pt,
    bottom=4pt,
    colback=blue!6!white,
    colframe=black,
    colbacktitle=black,
    enhanced,
    center,
    attach boxed title to top left={yshift=-0.1in,xshift=0.15in},
    boxed title style={boxrule=0pt,colframe=white,},
  }
}

\newtcolorbox{AIbox}[2][]{aibox,title=#2,#1}

\tcbset{
  takeawaybox/.style={
    width=\linewidth,
    top=8pt,
    bottom=4pt,
    colback=hidden-yellow,
    colframe=black,
    colbacktitle=black,
    enhanced,
    center,
    attach boxed title to top left={yshift=-0.1in,xshift=0.15in},
    boxed title style={boxrule=0pt,colframe=white,},
  }
}

\newtcolorbox{TakeawayBox}[2][]{takeawaybox,title=#2,#1}

\title{Beyond Surface Reasoning: Unveiling the True Long Chain-of-Thought Capacity of Diffusion Large Language Models}

\author{
  Qiguang Chen$^{1*}$ \quad Hanjing Li$^{1*}$ \quad Libo Qin$^{2, \coloremojicode{2709}}$ \quad Dengyun Peng$^1$ \quad Jinhao Liu$^1$ \quad Jiangyi Wang$^1$  \quad Chengyue Wu$^3$ \quad Xie Chen$^4$ \quad Yantao Du$^5$ \quad Wanxiang Che$^{1, \coloremojicode{2709}}$ \\
\normalfont{$^1$ LARG, Research Center for Social Computing and Interactive Robotics, Harbin Institute of Technology,\vspace{-5pt}\\
$^2$ School of Computer Science and Engineering, Central South University,\vspace{-5pt}\\
$^3$ The University of Hong Kong, \ \ $^4$ Shanghai Jiao Tong University,\ \ 
$^5$ ByteDance Seed (China) \\
}}

\begin{document}

\begin{abstract}
  \vspace{5mm}
  \textbf{\large Abstract:}
  \vspace{2mm}

 Recently, Diffusion Large Language Models (DLLMs) have offered high throughput and effective sequential reasoning, making them a competitive alternative to autoregressive LLMs (ALLMs). However, parallel decoding, which enables simultaneous token updates, conflicts with the causal order often required for rigorous reasoning. We first identify this conflict as the core Parallel–Sequential Contradiction (PSC).
 Behavioral analyses in both simple and complex reasoning tasks show that DLLMs exhibit genuine parallelism only for directly decidable outputs. As task difficulty increases, they revert to autoregressive-like behavior, a limitation exacerbated by autoregressive prompting, which nearly doubles the number of decoding steps with remasking without improving quality. Moreover, PSC restricts DLLMs' self-reflection, reasoning depth, and exploratory breadth. To further characterize PSC, we introduce three scaling dimensions for DLLMs: parallel, diffusion, and sequential. Empirically, while parallel scaling yields consistent improvements, diffusion and sequential scaling are constrained by PSC. Based on these findings, we propose several practical mitigations, parallel-oriented prompting, diffusion early stopping, and parallel scaling, to reduce PSC-induced ineffectiveness and inefficiencies.
\vspace{5mm}

  $^{*}$ \textit{Equal Contribution}
  
  $^{\coloremojicode{2709}}$ \textit{Corresponding Author}

  \vspace{5mm}

  \coloremojicode{1F4C5} \textbf{Date}: Oct 11, 2025

  \coloremojicode{1F4E7} \textbf{Contact}: \href{mailto:qgchen@ir.hit.edu.cn}{qgchen@ir.hit.edu.cn}, \href{mailto:car@ir.hit.edu.cn}{car@ir.hit.edu.cn}, \href{mailto:lbqin@csu.edu.cn}{lbqin@csu.edu.cn}

\end{abstract}
\maketitle

\vspace{3mm}
\pagestyle{headstyle}
\thispagestyle{empty}

\vspace{-2mm}\section{Introduction}\vspace{-1mm}

In recent years, diffusion large language models (DLLMs) have emerged as a novel generative paradigm, attracting increasing research 
attention~\citep{li2025survey,yang2025mmada}. Representative works such as LLaDA~\citep{nie2025llada} and Dream~\citep{ye2025dream} adopt a two-stage mask-denoising training strategy combined with parallel decoding for masked token prediction, effectively mitigating the ``reversal curse'' in traditional autoregressive large language models (ALLMs). Mercury~\citep{inception2025mercury} and Fast-DLLM~\citep{wu2025fast} further demonstrate the parallel efficiency of DLLMs, achieving an impressive generation speed in code tasks.

\begin{figure}[t]
    \centering
    \includegraphics[width=\textwidth]{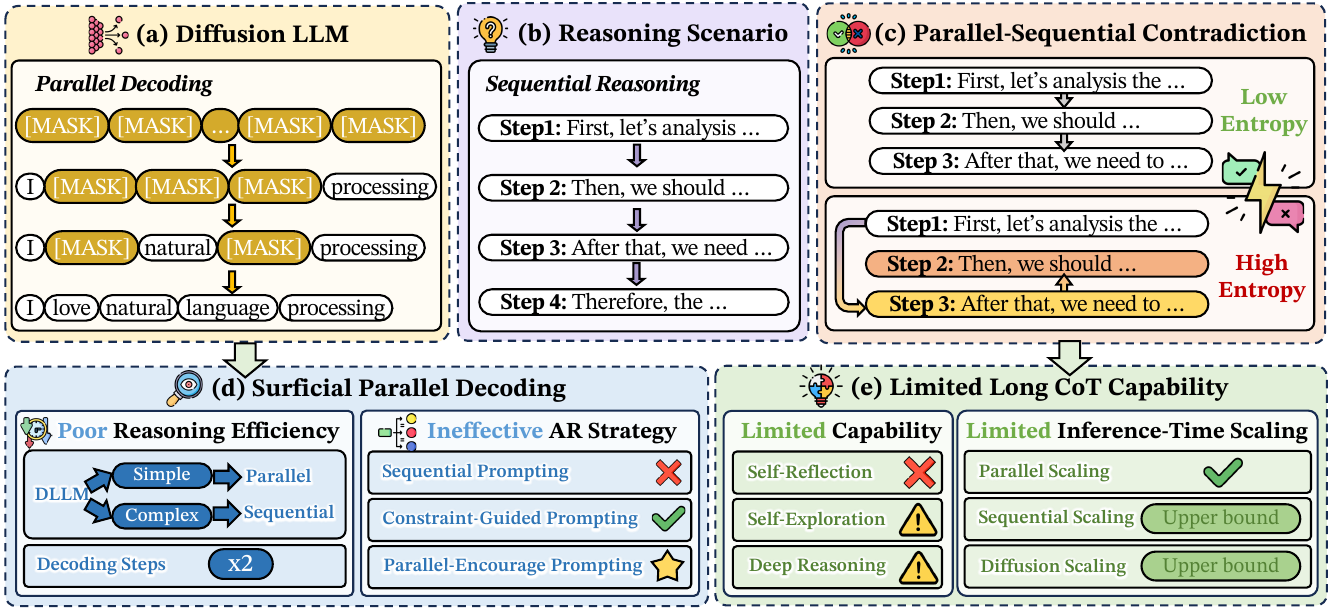}
    \caption{Overview of our work. Applying DLLMs to reasoning scenarios reveals an inherent contradiction between parallel processing and sequential reasoning, leading to high entropy, superficial parallel decoding, and limited Long CoT capabilities.\vspace{-8pt}}
    \label{fig:intro}
\end{figure}

Meanwhile, the rapid development of the Long Chain-of-Thought (Long CoT)~\citep{guo2025deepseek,chen2024unlocking,chen2025towards} has spurred increasing research on applying DLLMs to extended reasoning tasks~\citep{wang2025trado,wang2025diffusionLLMsD2F}.
\citet{zhao2025d1} and \citet{tang2025wd1} employ diffusion-augmented SFT and GRPO to further improve reasoning~\citep{gong2025diffucoder}. Moreover, Trado~\citep{wang2025trado} exploits overlooked information in sampling trajectories, achieving substantial gains.

As shown in Figure~\ref{fig:intro} (a), DLLMs generate text in parallel, producing a few non-sequential words in a single diffusion step. In sequential reasoning scenarios (Figure~\ref{fig:intro} (b)), the generation of step$_{i}$ requires the completion of step$_{i-1}$, leading to lower entropy~\citep{cui2025entropy,agarwal2025unreasonable}. In contrast, Figure~\ref{fig:intro} (c) shows DLLMs to parallel-decode by generating step$_{i+1}$ before step$_{i}$, resulting in high entropy.
Nevertheless, these parallel and sequential processes are inherently contradictory: \textbf{parallelism involves simultaneous processing, while sequential reasoning requires ordered steps.} To address this, we introduce the \textit{\textbf{Parallel–Sequential Contradiction (PSC)}}, which explores the underlying mechanisms and practical implications of diffusion-based reasoning.

To investigate this issue systematically, as shown in Figure~\ref{fig:intro} (d, e), we focus on two central research questions:
(1) \textbf{Do DLLMs truly perform parallel reasoning that avoids PSC?}
(2) \textbf{What challenges do DLLMs meet in Long CoT based on PSC?}
To address the first question, we analyze the decoding behavior of DLLMs in both simple and complex reasoning scenarios. Our findings show that DLLMs fail to achieve genuine parallel reasoning due to the PSC. They perform superficial parallel computation when outputs can be directly produced, but revert to an autoregressive mode under higher reasoning demands. This reliance on autoregression affects computational efficiency, which nearly doubles the computational cost with low confidence remasking. Furthermore, while autoregressive prompting is effective in ALLMs, it conflicts with DLLMs' parallel decoding design, amplifying the PSC of DLLMs. In contrast, strategies that reduce contradiction, such as conditional prompting or prompts that  encourage parallel generation, effectively enhance prompting performance.

To understand the second question, we examine the core capabilities of Long CoT in DLLMs. Our analysis reveals that, when faced with PSC, DLLMs often demonstrate limited self-reflection, shallow reasoning depth, and constrained exploratory behavior. Furthermore, we introduce three scaling dimensions for inference time, specifically designed for DLLMs: parallel, diffusion, and sequential scaling. Our findings show that both diffusion and sequential scaling are significantly constrained by PSC, while the parallel scaling law remains unaffected due to its vertical relationship with PSC.

In summary, our key contributions are as follows:
\begin{itemize}[leftmargin=16pt, itemsep=0pt, topsep=0pt]
    \item \textbf{Identification of Parallel-Sequential Contradictions}: To our knowledge, we first identify the Parallel-Sequential Contradiction (PSC) in DLLMs for Long CoT. We demonstrate that PSC leads to superficial parallel reasoning and reduced efficiency, requiring twice the decoding steps.
    \item \textbf{Systematic Exploration of DLLM Reasoning Limitation}: We conduct a systematic evaluation of DLLM reasoning, identifying the degradation of three core Long CoT capabilities, confirming the ineffectiveness of traditional autoregressive prompting methods, and demonstrating that diffusion scaling and sequential scaling are upper-bounded by PSC limitations.
    \item \textbf{Novel Mitigation Strategies}: We propose novel strategies to mitigate these issues and enhance DLLM reasoning. Our methods include parallel-encouraging prompting, diffusion early stopping, and parallel scaling, which substantially alleviate the constraints imposed by PSC.
\end{itemize}

\vspace{-2mm}\section{Parallel-Sequential Contradiction}\vspace{-1mm}
\label{sec:definition}

\subsection{Parallel Masked Diffusion Language Models}\vspace{-1mm}
In Diffusion Large Language Models (DLLMs), inference reconstructs missing spans by predicting masked tokens conditioned on a partially masked input. Its goal is modelling the conditional likelihood $p_\theta(x_0^i | x_l)$ for masked positions:
\begin{equation}
    -\mathbb{E}_{l, x_0, x_l} \left[
\frac{L}{l} \sum_{i=1}^L \mathbf{1}[x^i_l \in \mathcal{M}] \log p_\theta(x_0^i | x_l)
\right],
\end{equation}
where $L$ denotes the total number of tokens; $l$ is the number of masked tokens, uniformly sampled from $\{1, 2, \ldots, L\}$; $x_0$ is the complete original sequence.
$x_l$ is the partially masked sequence obtained by replacing those $l$ positions in $x_0$ with mask tokens, which serves as the conditional input. The indicator
$\mathbf{1}[x^i_l \in \mathcal{M}]$ equals $1$ if position $i$ is masked and $0$ otherwise.

\begin{figure}[t]
    \centering
    \includegraphics[width=\textwidth]{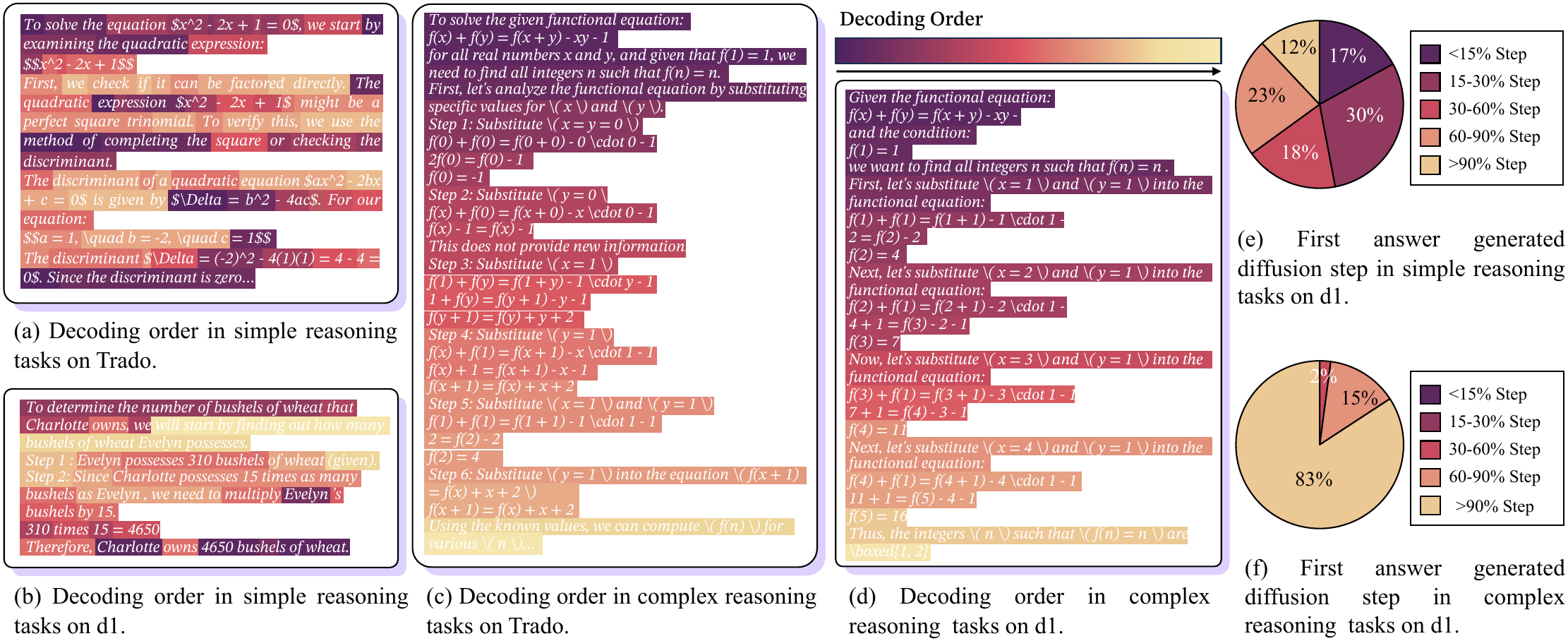}
    \caption{Diffusion order analysis with d1~\citep{zhao2025d1} and Trado~\citep{wang2025trado}, where later decoding orders are indicated by shallower colors.
    }
    \label{fig:diffusion-order}
\end{figure}

\vspace{-2mm}\subsection{Sequential Long Chain-of-Thought Reasoning}\vspace{-1mm}

Long Chain-of-Thought (Long CoT) allows LLMs to tackle complex problems by generating a sequence of reasoning steps. This method solves a problem \( P \) by following an ordered series of steps \( S_1, S_2, \ldots, S_n \), leading to the final answer \( A \).
Formally, it can be defined as:
\begin{equation}
p_{\theta}(A|P)=\prod_{t=1}^{n+1}p_{\theta}(S_t|P,S_{<t}).
\end{equation}
Here, $ S_{n+1}=A $, meaning the final answer is treated as the last step of the reasoning sequence. When generating each step $ S_t $, the model computes the conditional probability based on the problem $ P $ and all previously generated steps $ S_{<t} $.

\vspace{-2mm}\subsection{Parallel-Sequential Contradiction}\vspace{-1mm}
For \textbf{tasks with high parallelism}, downstream states typically yield predictable, high-probability outcomes, resulting in low predictive entropy. In these cases, optimizing the conditional probability \(p_\theta(S_k\mid S_1)\) is efficient, making non-autoregressive or semi-parallel generation methods advantageous.
In contrast, \textbf{tasks with strong sequential dependencies} exhibit high entropy when predicting distant future states in parallel. This uncertainty leads to significant predictive loss. To reduce this loss, the model is encouraged to break down the generation process into a sequence of low-entropy, step-by-step predictions. As a result, parallel generation conflicts with the model’s objective of identifying a low-loss, high-probability sequential path.
The formal proof is provided in Appendix~\ref{append:proof}.
\vspace{-2mm}\section{Do DLLMs truely perform parallel reasoning that avoids PSC?}\vspace{-1mm}
\subsection{Parallel-Sequential Contradictions cause superficial parallel reasoning.}\vspace{-1mm}
To examine whether DLLMs can genuinely perform parallel reasoning, we analyze their decoding behavior in both simple and complex scenarios. We have two key observations:\vspace{-16pt}

\paragraph{Parallel-Sequential Contradictions cause superficial parallel reasoning in simple scenarios.}
As shown in Figure~\ref{fig:diffusion-order} (a, b), DLLMs demonstrate parallel reasoning in simple cases where the model can direct output results without reasoning.
For instance, when solving "$x^2 + 2x + 1 = 0$", the model may simultaneously generate the Quadratic Formula "$\Delta = b^2 - 4ac$" and the final solution "$x = -1$" within a few diffusion steps. Following this, DLLMs complete the remaining reasoning steps in parallel, demonstrating the ability to leverage diffusion-based decoding to arrive at direct solutions without relying heavily on sequential reasoning.
To further explore this, we analyze the distribution of answers generated in the initial diffusion steps. As shown in Figure~\ref{fig:diffusion-order} (e), over 47\% of answers are produced within the first 30\% of diffusion steps. This suggests that in simple scenarios, DLLMs are capable of performing parallel reasoning, even though the underlying thought process, such as applying the Quadratic Formula before deriving the result, is inherently sequential.
\vspace{-16pt}

\paragraph{For complex reasoning, DLLMs converge toward autoregressive-like behavior to avoid PSC.}
To examine DLLM behavior in complex reasoning tasks, we validate PSC where the model cannot directly output the correct answer. Figure~\ref{fig:diffusion-order} (c, d) shows that DLLMs increasingly resemble autoregressive models. For example, when addressing tasks beyond direct generation, the model defaults to an autoregressive process. This suggests difficulty in sustaining parallel reasoning, which shifts to step-by-step processing. Figure~\ref{fig:diffusion-order} (f) further confirms this observation: in complex tasks, answers emerge later in the diffusion steps, reflecting a stronger reliance on ordered reasoning. These findings indicate that DLLMs face inherent PSC challenges in balancing parallel generation with sequential reasoning, ultimately converging toward autoregressive-like processing in complex scenarios.

\vspace{-2mm}\subsection{Diffusion-Step Dilemma: Sacrificing Efficiency Under PSC}\vspace{-1mm}
To investigate the reasoning efficiency of current DLLMs, we systematically categorize questions in BigGSM~\citep{chen2024unlocking} into different sampling lengths and diffusion steps (with low-confidence remasking). We evaluate two representative DLLMs under exponentially increasing diffusion steps and max token lengths (ranging from 1 to 1024). See Appendix~\ref{append:diffusion} for more details.\vspace{-16pt}

\begin{figure}[t]
    \centering
    \includegraphics[width=\textwidth]{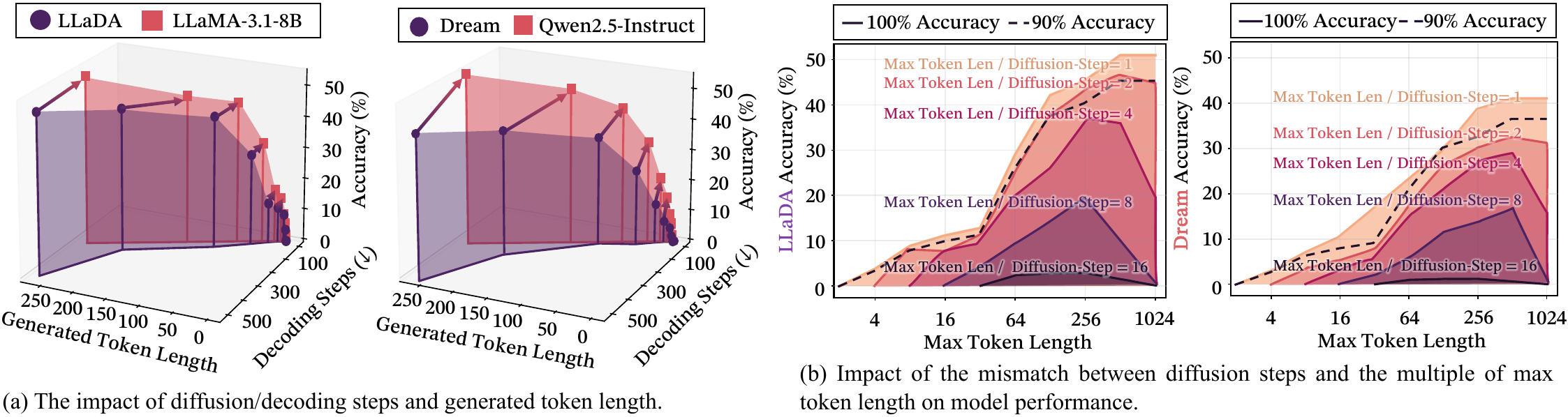}
    \caption{Diffusion speed analysis in Long-CoT-needed tasks with LLaDA-8B-Instruct~\citep{nie2025large} and Dream-7B-Instruct~\citep{ye2025dream} on BigGSM benchmark~\citep{chen2024unlocking}.\vspace{-16pt}}
    \label{fig:diffusion-speed}
\end{figure}

\paragraph{When complex reasoning, DLLMs require significantly more diffusion steps than ALLMs.} As shown in Figure~\ref{fig:diffusion-speed} (a), achieving comparable length and accuracy to ALLMs demands over 25\% more diffusion or decoding steps with remasking. In extreme cases, DLLMs require up to twice the token length in diffusion steps to match the performance and output length of autoregressive models when generating over 256 tokens. It indicates that effective reasoning entails roughly double the diffusion steps relative to the answer length, underscoring a notable efficiency challenge in reasoning tasks.\vspace{-16pt}

\begin{table*}[t!]
\centering
\resizebox{0.9\textwidth}{!}{
\begin{tabular}{l ccccc}
\toprule
\textbf{Model Name} & \textbf{BigGSM (Acc.)} & \textbf{GSM8K (Acc.)} & \textbf{Math-500 (Acc.)} & \textbf{HumanEval (Pass@1)} & \textbf{Average} \\

\midrule
Dream-7B-Instruct & 41.15 $_{(+0.00)}$ & 80.52 $_{(+0.00)}$ & 37.00 $_{(+0.00)}$ & 51.22 $_{(+0.00)}$ & 52.47 $_{(+0.00)}$ \\
\rowcolor[rgb]{0.942, 0.881, 0.998} \ \ +Zero-CoT & 41.15 $_{(-0.00)}$ & 77.26 $_{(-3.26)}$ & 34.00 $_{(-3.00)}$ & 48.78 $_{(-2.44)}$ & 50.30 $_{(-2.18)}$ \\
\rowcolor[rgb]{0.942, 0.881, 0.998}
\ \ +Plan-and-Solve & 34.75 $_{(-6.40)}$ & 78.85 $_{(-1.67)}$ & 20.20 $_{(-16.80)}$ & 48.78 $_{(-2.44)}$ & 45.65 $_{(-6.83)}$\\
\rowcolor[rgb]{0.942, 0.881, 0.998} \ \ +Least-to-Most & 35.25 $_{(-5.90)}$ & 77.03 $_{(-3.49)}$ & 16.20 $_{(-20.80)}$ & 43.29 $_{(-7.93)}$ & 42.94 $_{(-9.53)}$ \\
\rowcolor[rgb]{0.928, 0.936, 0.997} \ \ +Complex-CoT & 41.80 $_{(+0.65)}$ & 80.95 $_{(+0.43)}$ & 37.40 $_{(+0.40)}$ & \textbf{52.44 $_{(+1.22)}$} & 53.15 $_{(+0.68)}$ \\
\rowcolor[rgb]{0.928, 0.936, 0.997} \ \ +MARP & 42.95 $_{(+1.80)}$ & 80.52 $_{(+0.00)}$ & 37.20 $_{(+0.20)}$ & 51.22 $_{(+0.00)}$ & 52.97 $_{(+0.50)}$ \\
\rowcolor[rgb]{0.919, 0.969, 0.938} \ \ +Diff-MARP & \textbf{47.21 $_{(+6.06)}$} & \textbf{82.64 $_{(+2.21)}$} & \textbf{43.60 $_{(+6.60)}$} & \textbf{52.44 $_{(+1.22)}$} & \textbf{56.47 $_{(+4.00)}$}\\
\midrule
LLaDA-8B-Instruct & 48.03 $_{(+0.00)}$ & 75.36 $_{(+0.00)}$ & 34.80 $_{(+0.00)}$ & 32.32 $_{(+0.00)}$ & 47.63 $_{(+0.00)}$\\
\rowcolor[rgb]{0.942, 0.881, 0.998}
\ \ + Zero-CoT & 35.57 $_{(-12.46)}$ & 73.46 $_{(-1.90)}$ & 32.40 $_{(-2.40)}$ & 28.05 $_{(-4.27)}$ & 42.37 $_{(-5.26)}$ \\
\rowcolor[rgb]{0.942, 0.881, 0.998}
\ \ + Plan-and-Solve & 31.64 $_{(-16.39)}$ & 72.33 $_{(-3.03)}$ & 29.00 $_{(-5.80)}$ & 27.44 $_{(-4.88)}$ & 40.10 $_{(-7.53)}$\\
\rowcolor[rgb]{0.942, 0.881, 0.998}
\ \ + Least-to-Most & 34.75 $_{(-13.28)}$ & 73.31 $_{(-2.05)}$ & 30.80 $_{(-4.00)}$ & 27.44 $_{(-4.88)}$ & 41.58 $_{(-6.05)}$\\
\rowcolor[rgb]{0.928, 0.936, 0.997}
\ \ + Complex-CoT & 48.03 $_{(+0.00)}$ & 76.50 $_{(+1.14)}$ & 36.20 $_{(+1.40)}$ & 36.59 $_{(-4.27)}$ & 49.33 $_{(+1.70)}$\\
\rowcolor[rgb]{0.928, 0.936, 0.997}
\ \ + MARP & 48.20 $_{(+0.17)}$ & 76.35 $_{(+0.99)}$ & 34.40 $_{(-0.40)}$ & 35.37 $_{(-3.05)}$ & 48.58 $_{(+0.95)}$\\
\rowcolor[rgb]{0.919, 0.969, 0.938} \ \ +Diff-MARP & \textbf{55.74 $_{(+7.71)}$} & \textbf{76.80 $_{(+1.44)}$} & \textbf{38.20 $_{(+3.40)}$} & \textbf{38.41 $_{(+6.09)}$} & \textbf{52.29 $_{(+4.66)}$}\\
\midrule
LLaDA-v1.5 & 41.80 $_{(+0.00)}$ & 74.98 $_{(+0.00)}$ & 38.00 $_{(+0.00)}$ & 36.59 $_{(+0.00)}$ & 47.84 $_{(+0.00)}$\\
\rowcolor[rgb]{0.942, 0.881, 0.998}
\ \ + Zero-CoT & 36.39 $_{(-5.41)}$ & 71.87 $_{(-3.11)}$ & 37.20 $_{(-0.80)}$ & 35.98 $_{(-0.61)}$ & 45.36 $_{(-2.48)}$ \\
\rowcolor[rgb]{0.942, 0.881, 0.998}
\ \ + Plan-and-Solve & 30.16 $_{(-11.54)}$ & 74.37 $_{(-0.61)}$ & 34.40 $_{(-3.60)}$ & 35.98 $_{(-0.61)}$ & 43.73 $_{(-4.12)}$\\
\rowcolor[rgb]{0.942, 0.881, 0.998}
\ \ + Least-to-Most & 35.90 $_{(-5.90)}$ & 73.69 $_{(-1.29)}$ & 34.60 $_{(-3.40)}$ & 31.71 $_{(-4.88)}$ & 43.98 $_{(-3.87)}$\\
\rowcolor[rgb]{0.928, 0.936, 0.997} 
\ \ + Complex-CoT & 50.16 $_{(+8.41)}$ & 75.51 $_{(+0.53)}$ & \textbf{39.40 $_{(+1.40)}$} & \textbf{39.02 $_{(+2.43)}$} & 51.04 $_{(+3.19)}$\\
\rowcolor[rgb]{0.928, 0.936, 0.997}
\ \ + MARP & 42.13 $_{(+0.33)}$ & 74.37 $_{(0.61)}$ & 38.20 $_{(+0.20)}$ & 37.20 $_{(+0.61)}$ & 47.98 $_{(+0.13)}$\\
\rowcolor[rgb]{0.919, 0.969, 0.938} \ \ +Diff-MARP & \textbf{54.49 $_{(+12.79)}$} & \textbf{76.50 $_{(+1.52)}$} & 42.80 $_{(+4.80)}$ & 38.41 $_{(+1.82)}$ & \textbf{53.08 $_{(+5.23)}$}\\

\midrule
LLaDOU-Math & 42.13 $_{(+0.00)}$ & 81.88 $_{(+0.00)}$ & 45.80 $_{(+0.00)}$ & 39.02 $_{(+0.00)}$ & 52.21 $_{(+0.00)}$   \\
\rowcolor[rgb]{0.942, 0.881, 0.998} \ \ +Zero-CoT & 38.52$_{(-3.61)}$ & 80.95$_{(-0.93)}$ & 45.80$_{(-0.00)}$ & 37.80$_{(-1.22)}$ & 50.77$_{(-1.44)}$\\
\rowcolor[rgb]{0.942, 0.881, 0.998}
\ \ +Plan-and-Solve & 40.82$_{(-1.31)}$ & 81.12$_{(-0.76)}$ & 43.20$_{(-2.60)}$ & 38.41$_{(-0.61)}$ & 50.89$_{(-1.32)}$\\
\rowcolor[rgb]{0.942, 0.881, 0.998} \ \ +Least-to-Most & 40.16$_{(-1.97)}$ & 79.08$_{(-2.80)}$ & 43.00$_{(-2.80)}$ & 36.59$_{(-2.43)}$ & 49.71$_{(-2.50)}$\\
\rowcolor[rgb]{0.928, 0.936, 0.997}
\ \ + Complex-CoT & 43.77$_{(+1.64)}$ & 83.70$_{(+1.82)}$ & 45.80$_{(+0.00)}$ & \textbf{42.07 $_{(+3.05)}$} & 52.47$_{(+0.26)}$\\
\rowcolor[rgb]{0.928, 0.936, 0.997}
\ \ + MARP & 41.15$_{(-0.98)}$ & 82.18$_{(+0.30)}$ & 45.60$_{(-0.20)}$ & 40.26$_{(+1.24)}$ & 52.30$_{(+0.09)}$\\
\rowcolor[rgb]{0.919, 0.969, 0.938} \ \ +Diff-MARP & \textbf{54.26 $_{(+12.13)}$} & \textbf{84.76 $_{(+2.88)}$} & \textbf{49.00 $_{(+3.20)}$} & 40.85$_{(+1.83)}$ & \textbf{57.22 $_{(+5.01)}$}\\

\bottomrule
\end{tabular}
}
\caption{Performance comparison across 4 benchmarks. \textbf{Bold} marks the best baseline score per metric. For each method, we report its most token-efficient variant. Here, ``\raisebox{1pt}{\colorbox{mypurple}{ \rule[-0.2ex]{0pt}{1.0ex} }}'': prompting strategies, ``\raisebox{1pt}{\colorbox{myblue}{ \rule[-0.2ex]{0pt}{1.0ex} }}'': offline strategies, ``\raisebox{1pt}{\colorbox{mygreen}{ \rule[-0.2ex]{0pt}{1.0ex} }}'': online strategies.}
\label{tab:main_results}
\end{table*}

\paragraph{In reasoning scenarios, a large number of diffusion steps for autoregressive reasoning is unavoidable for acceptable accuracy.} Each generated token requires a sufficient number of diffusion iterations to allow the model to reason effectively and produce high-quality outputs. As illustrated in Figure~\ref{fig:diffusion-speed} (b), performance sharply declines when diffusion steps fall below the target token length. For example, generating 80 tokens with a maximum length of 128 but only 64 diffusion steps results in over a 10\% accuracy drop; with 32 steps, accuracy decreases by about 40\%. This demonstrates that inadequate diffusion severely impairs reasoning, as the model lacks enough refinement iterations. Thus, diffusion steps should at least match the planned token length to maintain reasoning quality. Nonetheless, excessive diffusion can significantly reduce efficiency.

\vspace{-2mm}\subsection{Rethinking the prompting strategies in DLLMs from PSC perspective}\vspace{-1mm}

In general, traditional autoregressive inference methods are typically categorized into two types: pipeline-guided approaches and condition-following approaches (see Appendix~\ref{append:prompt} for further details). In this section, we will begin by reviewing the theoretical foundations and representative implementations of these two categories. We will then examine their practical limitations and challenges. Furthermore, we introduce a parallel-encouraging prompting to improve DLLM effectiveness.\vspace{-16pt}

\paragraph{Sequential Reasoning Prompting will enlarge PSC's negative impact for DLLMs.}
Sequential prompting strategies, which facilitate sequential reasoning, have been shown to significantly improve the performance of ALLMs on complex tasks. However, as indicated in {\colorbox{mypurple}{purple rows}} of Table~\ref{tab:main_results}, we observed a notable decline in performance as tasks required an increasing number of reasoning steps. We attribute this decline to the fact that sequential reasoning prompts exacerbate the negative impact of PSC, thereby impairing the reasoning performance in DLLMs.\vspace{-16pt}

\paragraph{Constraint-guided Reasoning Prompting enhances model performance by preventing the introduction of additional PSC.}
By incorporating explicit constraints into the reasoning process, constraint-guided prompting effectively narrows the model's search space, thereby preventing the emergence of additional PSC during the reasoning process in DLLMs. This focused approach results in more accurate and reliable solutions. As shown {\colorbox{myblue}{blue rows}} of in Table~\ref{tab:main_results}, methods based on this principle, such as Complex-CoT~\citep{fu2022complexity} and MARP~\citep{chen2024unlocking}, demonstrate superior reasoning capabilities in DLLMs compared to traditional sequential prompting methods.\vspace{-16pt}

\paragraph{Parallel-encouraging Prompting reduces the sequential feature so that it further improves performance.}
Parallel-encouraging prompting refers to the technique of presenting multiple related tasks or questions simultaneously. This approach reduces the impact of PSC and minimizes the sequential features in the prompting process. By encouraging the model to make connections across these tasks, as illustrated in {\colorbox{mygreen}{green rows}} of Table~\ref{tab:main_results}, it effectively fosters DLLMs' performance, leading to more efficient reasoning and information integration. Leveraging the parallel processing capabilities of DLLMs, this method has the potential to significantly enhance performance, particularly in complex reasoning tasks, by promoting more comprehensive and coherent solutions.

\begin{AIbox}{Takeaways}
\begin{enumerate}[leftmargin=*]
    \item Due to PSC, DLLMs engage in superficial parallel reasoning and exhibit autoregressive behavior in complex scenarios, which compromises their reasoning efficiency.
    \item Sequential prompts prove ineffective for DLLMs, requiring PSC-free or -redcued approaches like constraint-guided and parallel-encouraging prompts to guide their operation.
\end{enumerate}
\end{AIbox}

\vspace{-1mm}\section{What challenges do DLLMs meet in Long CoT based on PSC?}\vspace{-1mm}
Despite impressive empirical results, DLLMs' genuine reasoning abilities and scalability under Parallel-Sequential Contradictions remain open questions. We systematically evaluate Long CoT to assess these fundamental capabilities and scaling strategies.

\vspace{-2mm}\subsection{DLLMs do not have sufficient basic capabilities to support Long CoT.}\vspace{-1mm}

Long CoT is the primary innovation in recent reasoning large language models, leveraging inference-time scaling for self-exploration, self-reflection, and deep reasoning~\citep{chen2025towards}. Evaluation details are in Appendix~\ref{append:capability}.\vspace{-16pt}

\begin{figure*}[t]
    \centering
    \includegraphics[width=\textwidth]{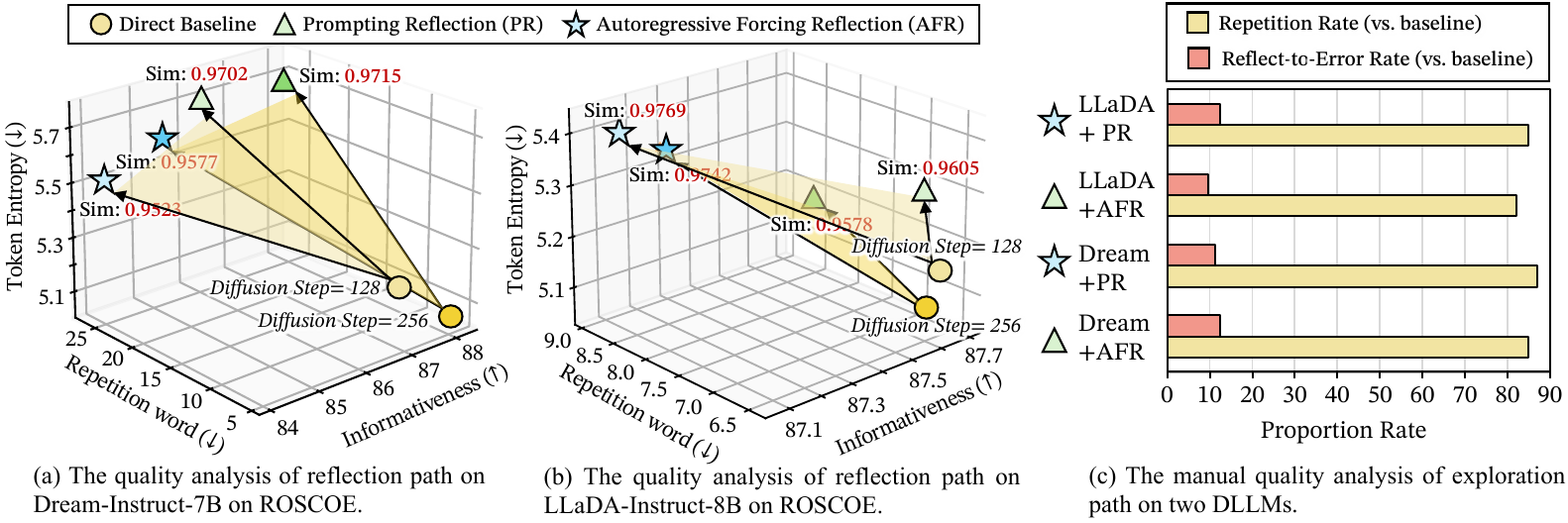}
    \caption{Self-reflection performance and rationale quality evaluation on DLLMs.}
    \label{fig:reflection}
\end{figure*}

\paragraph{Traditional reflection strategies are Ineffective for DLLMs.}
Long CoT models always employ a self-reflection mechanism for iterative reasoning refinement. To assess its efficacy, we examine two LLM paradigms: (1) Prompting Reflection and (2) Autoregressive Forcing Reflection
As shown in Figure~\ref{fig:reflection} (a, b), reflection paradigms yield no significant differences from vanilla reasoning chains in semantic similarity, informativeness, or token-level entropy. Though the reflection process increases entropy and reduces informativeness, it maintains over 0.95 semantic similarity to original reasoning chains.
These findings suggest the reflection mechanism offers only limited surface-level optimization. Figure~\ref{fig:reflection} (c) further reveals a substantial token repetition ratio compared to the original path, resulting in approximately 10\% reflection-to-error responses.\vspace{-16pt}

\begin{figure*}[t]
    \centering
    \includegraphics[width=\textwidth]{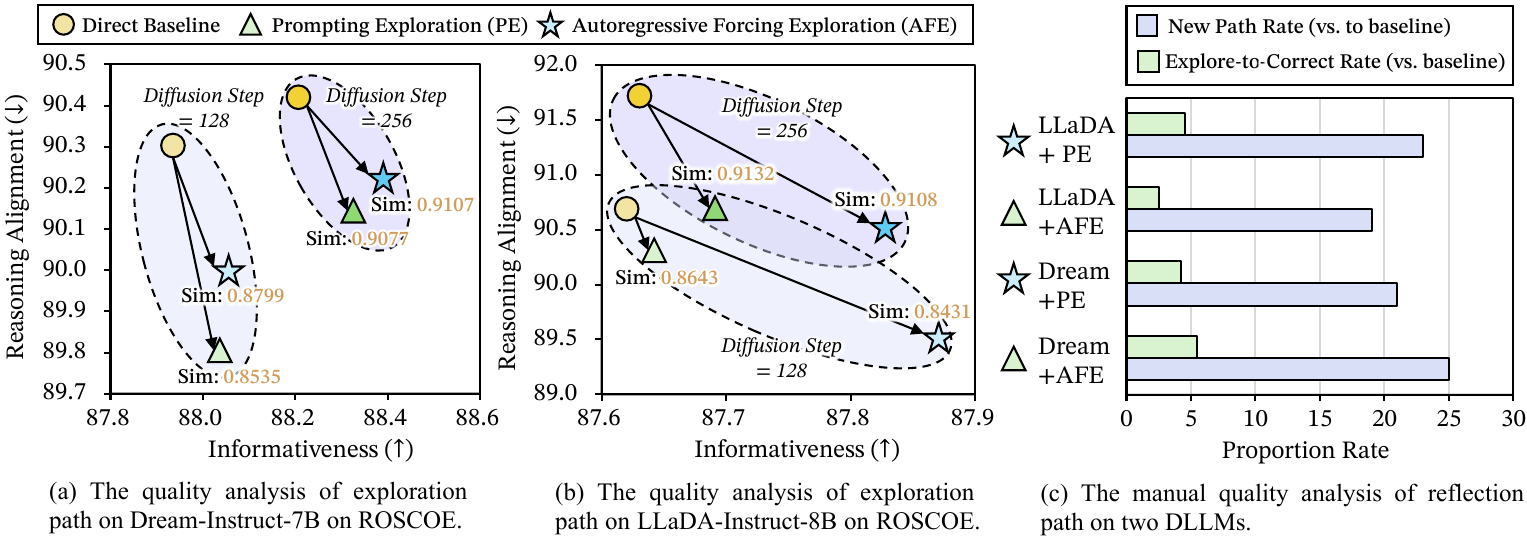}
    \caption{Self-exploration performance and rationale quality evaluation on DLLMs.}
    \label{fig:exploration}
\end{figure*}

\paragraph{Limited Efficacy of traditional exploration strategies for novel reasoning path generation.}
Exploration, a fundamental competency for complex reasoning, involves a model's ability to generate diverse and innovative solutions. To assess this potential in DLLMs, we designed experiments utilizing two strategies: (1) Prompting Exploration and (2) Autoregressive Forcing Exploration. Figure~\ref{fig:exploration} (a, b) reveal that current exploration strategies offer several improvements in the novel semantic of generated reasoning processes. However, these improvements remain superficial, evidenced by a high similarity (> 0.84) between explored paths and original results. Furthermore, as depicted in Figure~\ref{fig:exploration} (c), while the new path and explore-to-correct ratios are limited ($\thicksim$ 5\%), they nonetheless indicate a positive, albeit constrained, effect.\vspace{-16pt}

\begin{figure}[t]
    \centering
    \includegraphics[width=\textwidth]{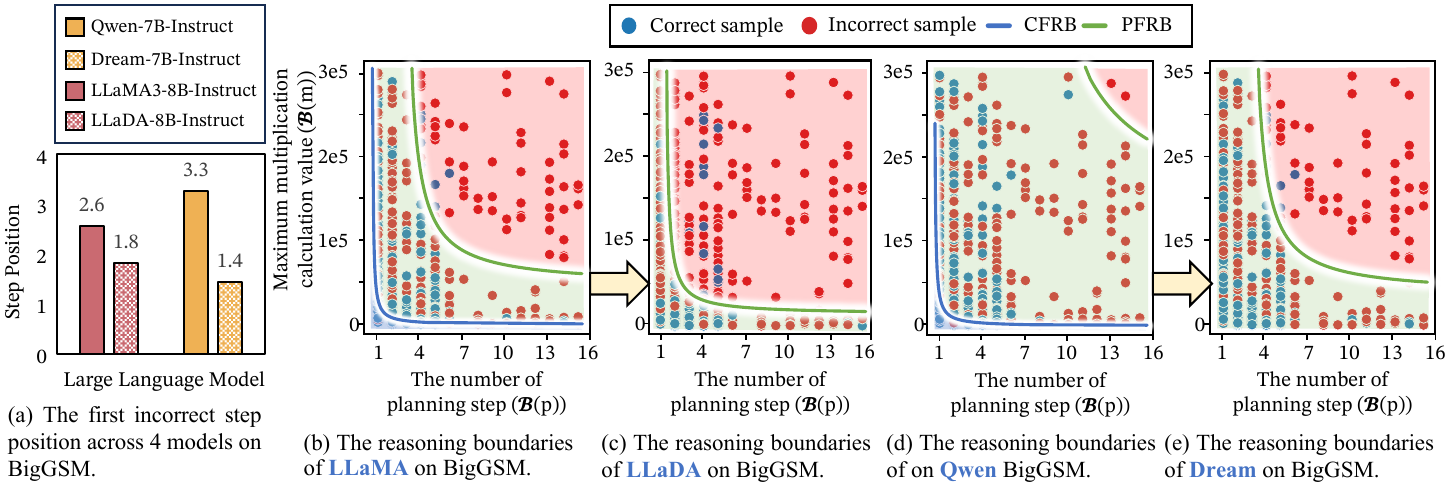}
    \caption{Incorrect Step and Reasoning Boundaries Distribution of DLLMs on BigGSM.}
    \label{fig:bound}
\end{figure}

\paragraph{DLLMs possess limited reasoning boundaries and, consequently, exhibit restricted deep reasoning abilities.}
To examine the limitations of DLLMs on deep reasoning, we evaluate their capacity to sustain reasoning across sufficient depths. Figure~\ref{fig:bound} (a) demonstrates that error steps are all less than 2, which suggests that current DLLMs are unable to consistently sustain deep reasoning performance. Furthermore, following \citet{chen2024unlocking}, we define the 90\% correctness step count as the models’ completely feasible reasoning boundary (CFRB), and the 10\% correctness step count as the completely infeasible reasoning boundary (CIRB). As shown in Figure~\ref{fig:bound} (b), current DLLMs display similar CFRB values but lower CIRB values, indicating narrower feasible reasoning ranges.

\vspace{-2mm}\subsection{Current DLLMs have three-directional but limited Inference-Time Scaling}\vspace{-1mm}
Given their denoising characteristics, we investigate a fundamental question: \textbf{Is there also Inference-Time Scaling Law in DLLM under such contradictions?} We examine this through three complementary perspectives: Parallel Scaling, Diffusion Scaling, and Sequential Scaling. These experiments determine whether DLLMs follow inference-time scaling laws and provide practical insights for optimizing reasoning performance. Implementation details can be seen in Appendix~\ref{append:scaling}.

\begin{figure*}[t]
    \centering
    \begin{minipage}{0.42\textwidth}
        \centering
        \includegraphics[width=\textwidth]{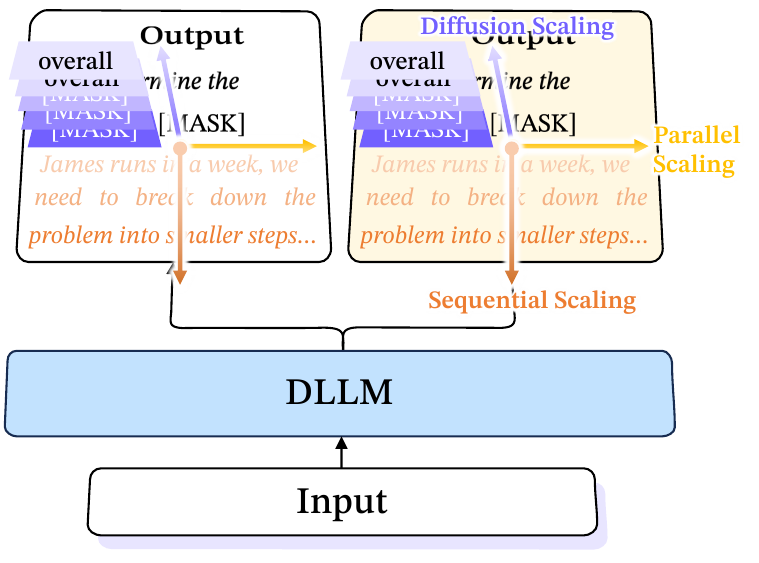}
        \caption{Three primary scaling directions for DLLMs proposed in our work.}
        \label{fig:scaling}
    \end{minipage}
    \hfill
    \begin{minipage}{0.54\textwidth}
        \centering
        \includegraphics[width=\textwidth]{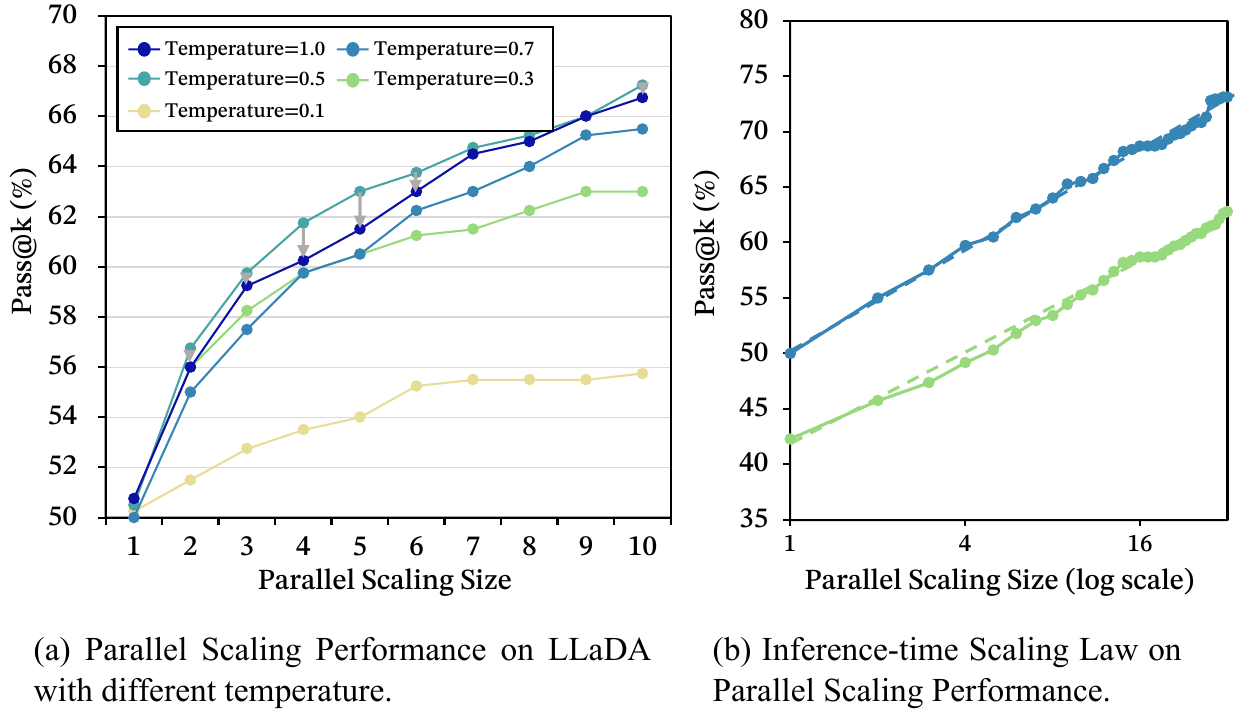}
        \caption{Performance analysis under Parallel Scaling.}
        \label{fig:parallel-scaling}
    \end{minipage}
\end{figure*}

\vspace{-2mm}\subsubsection{Parallel Scaling Law holds Despite PSC}\vspace{-1mm}
For DLLMs, a key question is whether their unique diffusion generation mechanism supports efficient parallel sampling and whether parallel sampling can effectively enhance reasoning performance.\vspace{-16pt}

\paragraph{Higher temperatures do not always yield more diverse and effective parallel sampling.} 
The decoding temperature controls generation randomness, with higher values typically increasing output diversity in ALLM reasoning. We adjust the temperature during generation (0.1 to 1.0) to evaluate its impact on parallel sampling.
Model accuracy improves steadily with increasing \texttt{Pass@k} values across all temperature settings before plateauing. As shown in Figure~\ref{fig:parallel-scaling} (a), moderate temperatures (e.g., $T=0.5$) achieve optimal performance, while both lower and higher temperatures yield diminished performance gains. This indicates that moderate temperature settings provide the optimal balance between generation diversity and output reliability.\vspace{-16pt}

\paragraph{DLLM reasoning accuracy improves with increased parallel samples, following inference-time scaling patterns.}
As shown in Figure~\ref{fig:parallel-scaling} (b), when $k$ increases from 1 to 32, accuracy demonstrates nearly linear improvement on a logarithmic scale. This indicates that DLLMs effectively utilize parallel sampling to enhance reasoning performance, as diverse outputs increase the probability of generating correct solutions. This pattern aligns with inference-time scaling laws observed in other advanced language models, where performance scales with computational effort during inference.

\begin{figure}[t]
    \centering
    \includegraphics[width=0.96\textwidth]{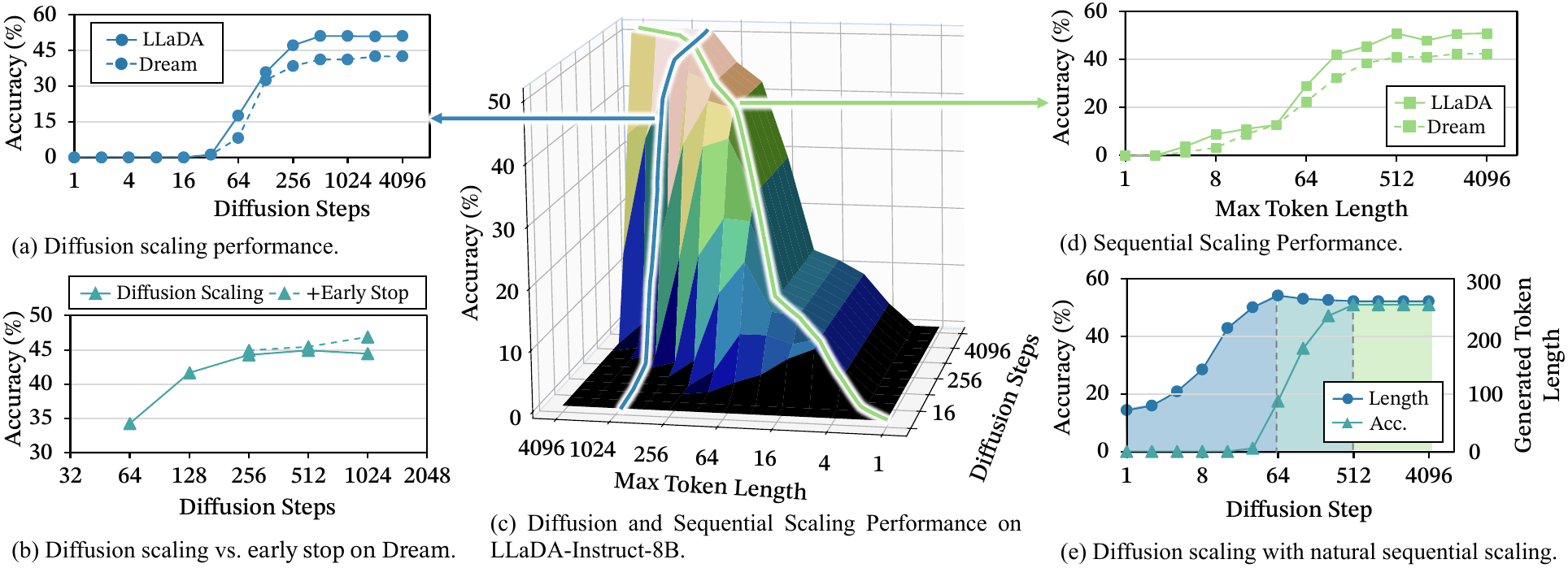}
    \caption{Diffusion Scaling analysis of reasoning accuracy across difficulty levels on  BigGSM.}
    \label{fig:diffusion-scaling}
\end{figure}

\vspace{-2mm}\subsubsection{Diffusion Scaling Law is Broken and Constrained by PSC}\vspace{-1mm}
\paragraph{Diffusion Scaling of DLLMs ensures performance gains, with diffusion time positively correlated.} Model accuracy increases monotonically with the number of diffusion steps. We are the first to formalize \emph{Diffusion Scaling} in DLLMs, proposing a positive correlation between model performance and diffusion iterations.
To validate this claim, we benchmark two representative DLLMs, DREAM~\citep{ye2025dream} and LLaDA~\citep{nie2025large}, under an exponential schedule of diffusion steps (1–1024). By tracking accuracy at each step, we observe how DLLMs address reasoning tasks of varying complexity across the diffusion process. As shown in Figure~\ref{fig:diffusion-scaling} (a), performance consistently improves with deeper diffusion; however, the rate of improvement depends on task difficulty: simpler problems gain substantially, while tasks beyond the model’s capacity yield only limited benefits.\vspace{-16pt}

\paragraph{Diffusion Scaling is effective and exhibits an upper bound, beyond which an over-diffusion phenomenon emerges due to PSC.}
Consistent with classical scaling laws, the benefits of diffusion scaling are inherently capped. As shown in Figure~\ref{fig:diffusion-scaling} (a), increasing diffusion steps improves performance from 32 to 512 steps, after which gains plateau. More importantly, excessive diffusion reduces accuracy: Figure \ref{fig:diffusion-scaling} (b) shows a drop from 44.92\% to 44.43\%. This decline illustrates \emph{over-diffusion}, where extended denoising introduces excessive corrections that disrupt reasoning chains, akin to overfitting caused by training without early stopping.\vspace{-16pt}

\paragraph{Early stopping can effectively mitigate over-diffusion.}
To address over-diffusion, we propose a Diffusion Early Stopping (DES) strategy that halts the process when generated tokens stabilize. The implementation comprises three components: (1) \textbf{Overlap Ratio Calculation}: computed as the proportion of identical tokens between consecutive steps. (2) \textbf{Convergence Detection}: potential convergence occurs when the overlap ratio meets or exceeds a predefined threshold. (3) \textbf{Activation Condition}: early stopping triggers only after three consecutive steps satisfy the threshold, preventing false positives from transient fluctuations.
As shown in Figure~\ref{fig:diffusion-scaling} (b), we observe that beyond 256 steps, early stopping outperforms standard diffusion, with accuracy improving from 44.26\% to 46.89\% at 1024 steps. Early stopping captures convergence states and terminates upon stabilization, while preventing performance degradation from excessive diffusion.

\vspace{-2mm}\subsubsection{Sequential Scaling Law is also Broken and Constrained by PSC}\vspace{-1mm}
\paragraph{The inherent limitations of sequential scaling for DLLMs.} While sequential scaling has shown promise in enhancing reasoning capabilities, it remains constrained by the inherent characteristics of DLLMs. As shown in Figure~\ref{fig:diffusion-scaling} (d), the performance improvements are at first increasing but eventually converge. This limitation arises from the fact that sequential scaling relies on the model's ability to maintain context over extended reasoning chains, a challenge for current DLLMs.\vspace{-16pt}

\paragraph{Sequential Scaling also meets over-thinking challenges.} Similar to diffusion scaling, as shown in Figure~\ref{fig:diffusion-scaling} (d), sequential scaling faces its own set of over-thinking challenges. As the model attempts to extend its reasoning across longer contexts, it may encounter diminishing returns or even performance degradation. This phenomenon is particularly evident in tasks that require intricate reasoning over extended text, where the model's ability to track and integrate information can become strained. \vspace{-16pt}

\paragraph{Diffusion Scaling can naturally yield Sequential Scaling benefits.} As shown in Figure~\ref{fig:diffusion-scaling} (e), diffusion scaling alleviates the limitations of sequential scaling. We identify three stages in DLLMs during diffusion: (1) sequential scaling, (2) compression, and (3) convergence. In the first stage, increasing diffusion steps leads to stable performance but longer solutions, indicating that DLLMs explore suitable lengths for reasoning. In the second stage, the model compresses its reasoning by eliminating redundancy, generating more efficient solutions. In the third stage, the model converges on an optimal strategy, achieving high performance while reducing computational cost.

\begin{AIbox}{Takeaways}
\begin{enumerate}[leftmargin=*]
    \item DLLMs are deficient in three basic Long CoT capabilities, hindering their effectiveness.
    \item DLLM can be optimized via parallel, diffusion, and sequential scaling. Diffusion scaling inherently encompasses the benefits of sequential scaling.
    \item The performance of both diffusion and sequential scaling is ultimately upper-bounded by a parallel-sequential contradiction. But Parallel scaling law remains the most effective strategy, although it is also the most computationally expensive.
\end{enumerate}
\end{AIbox}

\vspace{-2mm}\section{Related work}\vspace{-1mm}

The application of diffusion models to text generation has emerged as an alternative to autoregressive methods. Early work by D3PM~\citep{austin2021structured} proposed discrete denoising diffusion probabilistic models, and Diffusion-BERT~\citep{he2022diffusionbert} demonstrated scalability to BERT-style architectures. SEDD~\citep{lou2023discrete} achieved performance comparable to GPT-2.
Recent progress has broadened the scope of Diffusion Large Language Models (DLLMs)~\citep{yang2025mmada,wu2025fast,gong2025diffucoder}. LLaDA~\citep{nie2025large} and Dream~\citep{ye2025dream} scaled to billion-parameter models with notable inference gains.
The D2F strategy~\citep{wang2025diffusionLLMsD2F} further enhanced inference by enabling block-level autoregression and parallel decoding, maintaining a balance between speed and accuracy. This direction aligns with the growing interest in applying DLLMs to extended reasoning~\citep{wang2025trado,zhao2025d1}.
Diffusion-of-Thought (DoT)~\citep{ye2024diffusion} combines diffusion with chain-of-thought reasoning. Building on this, \citet{zhao2025d1} and \citet{tang2025wd1} applied diffusion-augmented SFT and GRPO to strengthen reasoning. Similarly, Trado~\citep{wang2025trado} exploits overlooked sampling signals, yielding further reasoning gains.

However, while DLLMs exhibit notable parallel decoding in text generation and consistently strong step-by-step reasoning, these features appear conceptually opposed: parallelism implies simultaneous processing, whereas sequential reasoning demands ordered progression. This apparent Parallel–Sequential Contradiction (PSC) suggests that both the underlying mechanisms and the practical effectiveness of DLLMs’ diffusion-based reasoning remain insufficiently understood.

\vspace{-2mm}\section{Conclusion}\vspace{-1mm}
In this work, we formalize the Parallel-Sequential Contradiction (PSC) to explain why DLLMs, though built for parallel decoding, revert to autoregression as reasoning difficulty rises. Empirically, DLLMs exploit parallelism only when tokens are locally decidable; otherwise, they fall back to sequential computation, reducing efficiency. Further, we first define three-dimensional scaling: parallel, diffusion, and sequential scaling, and show that PSC restricts the latter two while parallel scaling holds. We mitigate PSC through parallel-focused prompting, diffusion early stopping, and parallel scaling, improving both accuracy and throughput. Future work should align training and architectures with PSC-aware reasoning and design benchmarks, isolating its effects.

\bibliographystyle{./refstyle}
\bibliography{ref}

\appendix
\newpage
\begin{center}
    \textbf{\LARGE Appendix}
\end{center}
  
\section{Mathematical Proof of DLLM Degrading to Autoregressive}
\label{append:proof}

\paragraph{Goal.}
We rigorously show, via information theory and optimization, that the intrinsic statistical property of a generative task, namely its sensitivity to perturbations of initial conditions, fundamentally determines its optimal (lowest-loss) generation strategy. 
Concretely, we axiomatize two classes of tasks: serial tasks (step-by-step reasoning) exhibiting cascading sensitivity to initial conditions, and parallel tasks exhibiting partial invariance, and we prove that serial tasks induce significantly higher conditional entropy for ``skip-step'' parallel predictions \(S_k \mid S_1\) than parallel tasks, forcing any loss-minimizing learner to converge to an autoregressive strategy. 

\subsection{Problem Setup and Notation}

Let all step states \(S_i\) take values in a metric space \((\Omega,d)\). 
Specifically, the true data distribution \(p_{\theta}^S\) of a task is considered serial if, for any given \(s_1 \in \Omega\), the mapping from \(s'_1\) to a subsequent state \(s'_k\) is highly divergent within a sufficiently small neighborhood \(N(s_1, \epsilon) = \{s'_1 | d(s_1, s'_1) < \epsilon\}\). Conversely, a task's true data distribution \(p_{\theta}^P\) is considered parallel if, for any given \(s_1 \in \Omega\), there exists at least one subsequent state \(S_k\) (where k>1) that is insensitive to perturbations within its neighborhood \(N(s_1, \epsilon)\). Formally, this leads to the following definitions for the two tasks.

\begin{definition}[Serial tasks: cascading sensitivity]
A data-generating distribution \(p_{\theta}^S\) is serial if for any \(s_1\in\Omega\) and any \(k>1\), that satisfies:
\begin{equation}
    \lim_{\varepsilon\to 0}\; \mathbb{E}_{s_1'\sim U(N(s_1,\varepsilon))}\big[p_{\theta}^S(S_k = s_k \mid S_1=s_1')\big] \;=\; 0,
\end{equation}
where \(s_k\) is the reference outcome drawn from the true conditional \(p_{\theta}^S(S_k\mid S_1=s_1)\), and \(U(N(s_1,\varepsilon))\) the uniform distribution over this neighborhood. 
Equivalently, arbitrarily small perturbations of \(S_1\) almost surely drive future states away from the reference trajectory at step \(k\), capturing sensitive dependence on initial conditions.
\label{def:serial}
\end{definition}

\begin{definition}[Parallel tasks: partial invariance]
A data-generating distribution \(p_{\theta}^P\) is parallel if there exists some \(k>1\) and a constant \(C \in (0,1]\) such that for any \(s_1\in\Omega\), that satisfies:
\begin{equation}
\lim_{\varepsilon\to 0}\; \mathbb{E}_{s_1'\sim U(N(s_1,\varepsilon))}\big[p_{\theta}^P(S_k = s_k \mid S_1=s_1')\big] \;=\; C \in (0,1],
\end{equation}
where \(s_k\) is the reference outcome drawn from \(p_{\theta}^P(S_k\mid S_1=s_1)\). 
Thus, a structurally stable downstream state persists with significant probability despite infinitesimal perturbations of the initial condition. 
\label{def:parallel}
\end{definition}

\subsection{Learning problem}

Let \(p_{\theta}\) be a parametric generative model trained by minimizing cross-entropy with respect to the true data distribution \(\hat{p}\), i.e.,
\begin{equation}
    L(p_{\theta}, \hat{p})=\mathbb{E}_{x\sim \hat{p}}[-\log p_{\theta}(x)] = H(\hat{p})+D_{\mathrm{KL}}(\hat{p}\Vert p_{\theta}),
\end{equation}
so minimizing cross-entropy is equivalent to minimizing \(D_{\mathrm{KL}}(\hat{p}\Vert p_{\theta})\) and to maximum likelihood. 
For any conditional subproblem, the optimum satisfies \(p^*_{\theta}(S_k\mid S_1)=\hat{p}(S_k\mid S_1)\), and the minimal expected negative log-likelihood equals the conditional entropy,
\begin{equation}
L^* =\mathbb{E}_{s_1\sim \hat{p}(S_1)}\Big[H\big(\hat{p}(S_k\mid S_1=s_1)\big)\Big],
\text{where } 
H(Y\mid X)\equiv -\!\!\!\sum_{x\in X, y\in Y}p(x,y)\log \frac{p(x,y)}{p(x)}.\vspace{-16pt}
\end{equation}

\paragraph{Autoregression and chain rule.}

A step-by-step reaosning strategy factorizes a joint distribution as a product of conditionals via the chain rule, thereby replacing a high-entropy ``skip'' conditional \(p_{\theta}(S_k\mid S_1)\) by a sequence of typically lower-entropy one-step conditionals \(p_{\theta}(S_{t+1}\mid S_t,\ldots)\). 
It is the standard rationale behind likelihood-based training of ALLMs under teacher forcing. 

\subsection{Discretization} 
To compare entropies on a general metric space, consider a finite measurable partition \(\Pi_\varepsilon\) of \(\Omega\) with mesh size at most \(\varepsilon\), and define the discretization and quantization map \(\phi_\varepsilon:\Omega\to[m_\varepsilon]\) that assigns each \(s\in\Omega\) to its cell index, where \(m_\varepsilon=|\Pi_\varepsilon|\). 
Let \(\tilde S_i^{(\varepsilon)}=\phi_\varepsilon(S_i)\) and write \(p_{\theta}^{(\varepsilon)}(\cdot)\) for the induced discrete laws; we analyze \(H(\tilde S_k^{(\varepsilon)}\mid \tilde S_1^{(\varepsilon)}=\tilde s_1)\), which is well-defined, and relate back to the original problem by taking \(\varepsilon\to 0\). 
Two standard facts underpin the analysis: (i) for a fixed finite support, entropy is maximized by the uniform distribution; (ii) the Shannon entropy is bounded below by the min-entropy \(-\log p_{\max}\), and admits tighter lower bounds in terms of the binary entropy function \(H_b\) and the support size.

\begin{lemma}[Pointwise probability caps]
Fix \(k>1\) and \(s_1\in\Omega\). Under Definition~\ref{def:serial}, for any \(\delta>0\) there exists \(\varepsilon_0>0\) such that for all \(\varepsilon<\varepsilon_0\),
\begin{equation}
\max_{s'_1 \sim N'(s_1)} p_{\theta,S}^{(\varepsilon)}\!\big(\tilde S_k^{(\varepsilon)}=s_k \mid \tilde S_1^{(\varepsilon)}=\phi_\varepsilon(s'_1)\big) \;\le\; \delta,
\end{equation}
where $s'_1 \sim N'(s_1)\Leftrightarrow \varepsilon(s_1) \neq \varepsilon(s'_1) \wedge  \varepsilon(s'_1) \in m_\varepsilon$.

Under Definition~\ref{def:parallel}, there exist \(C>0\) and \(\varepsilon_0>0\) such that for all \(\varepsilon<\varepsilon_0\),
\begin{equation}
\max_{s'_1 \sim N'(s_1)} \big(\tilde S_k^{(\varepsilon)}=s_k \mid \tilde S_1^{(\varepsilon)}=\phi_\varepsilon(s'_1)\big) \;\ge\; C.
\end{equation}
\end{lemma}

\noindent
\emph{Proof sketch.}
By Definition~\ref{def:serial}, for serial tasks, the conditional probability of a reference outcome, averaged over shrinking neighborhoods of \(s'_1\), vanishes. This forces any mass that could be concentrated on a particular cell containing \(s_k\) to diminish as the mesh refines. In contrast, for parallel tasks, Definition~\ref{def:parallel} guarantees a persistent mass \(C \in (0,1]\) associated with a stable outcome across neighborhoods, which uniformly lower-bounds the maximum conditional atom in \(\varepsilon\).

\subsection{Main proposition and quantitative bounds} 
\begin{proposition}[Skip-step parallel predictions on serial vs. parallel tasks]
For any \(k>1\), the optimal expected skip-prediction loss on serial-task data strictly exceeds that on parallel-task data:
\begin{equation}
L^*_S\big(p_{\theta}(S_k\mid S_1=s'_1)\big) \;>\; L^*_P\big(p_{\theta}(S_k\mid S_1=s'_1)\big),
\end{equation}
equivalently,
\begin{equation}
\mathbb{E}_{s_1'\sim U(N(s_1,\varepsilon))}\!\big[H_S\big(S_k\mid S_1=s'_1\big)\big] \;>\; \mathbb{E}_{s_1'\sim U(N(s_1,\varepsilon))}\!\big[H_P\big(S_k\mid S_1=s'_1\big)\big],
\end{equation}
In the discrete case, this reduces to showing
\begin{equation}
\mathbb{E}_{s'_1 \sim N'(s_1)}\!\big[H_S\big(\tilde S_k^{(\varepsilon)}\mid \tilde S_1^{(\varepsilon)}=\phi_\varepsilon(s_1)\big)\big]
>
\mathbb{E}_{s'_1 \sim N'(s_1)}\!\big[H_P\big(\tilde S_k^{(\varepsilon)}\mid \tilde S_1^{(\varepsilon)}=\phi_\varepsilon(s_1)\big)\big]
\end{equation}
with a strictly positive gap that can be quantified through discretization and classical entropy bounds.
\end{proposition}

\noindent
\emph{Proof.}
It suffices to compare the conditional entropies pointwise and then take expectations. Fix \(s_1\) and a partition \(\Pi_\varepsilon\). 
We first define the maximum of generation probability of serial tasks:
\begin{equation}
    p_{\max}^{S}(\varepsilon;s'_1) := \max_{s'_1 \sim N'(s_1)} p_{\theta,S}^{(\varepsilon)}\!\big(\tilde S_k^{(\varepsilon)}=s_k \mid \tilde S_1^{(\varepsilon)}=\phi_\varepsilon(s'_1)\big)
\end{equation}
and analogously \(p_{\max}^{P}(\varepsilon;s'_1)\) under $p_{\theta,P}^{(\varepsilon)}$.

By the lemma, \(p_{\max}^{S}(\varepsilon;s'_1)\to 0\) as \(\varepsilon\to 0\), while for parallel tasks one has \(p_{\max}^{P}(\varepsilon;s'_1)\ge C\) for all sufficiently small \(\varepsilon\). 
For any discrete distribution over \(m_\varepsilon\) points with maximal atom \(p_{\max}\), Fano’s inequality implies 
\begin{equation}
    H \le H_b(p_{\max}) + (1-p_{\max})\log(m_\varepsilon-1),
\end{equation}
where $H_b$ is the binary entropy.

Thus, for parallel tasks,
\begin{equation}
H_P\big(\tilde S_k^{(\varepsilon)}\mid \tilde S_1^{(\varepsilon)}=\phi_\varepsilon(s'_1)\big)
\;\le\; H_b\!\big(p_{\max}^{P}(\varepsilon;s'_1)\big) + \big(1-p_{\max}^{P}(\varepsilon;s'_1)\big)\log(m_\varepsilon-1),
\end{equation}
and since \(p_{\max}^{P}(\varepsilon;s'_1)\ge C>0\), the entropy is uniformly bounded away from the maximal value \(\log(m_\varepsilon)\) by a constant determined by \(C\). 

For serial tasks, since \(p_{\max}^{S}(\varepsilon;s'_1)\to 0\), we have error probability $p_e \rightarrow 0$. Now, we should apply the contrapositive of Fano’s inequality. Specifically, given the Fano's inequality:
\begin{equation}
    H \le H_b(p_e) + p_e\log(m_\varepsilon-1), 
\end{equation}
it follows that \(H \rightarrow 0 \Rightarrow p_e \rightarrow 0\). Conversely, $p_e \rightarrow 1$ implies  $H \rightarrow H_{max}$. In this sense, the condition is satisfied:
\begin{equation}
H_P\big(\tilde S_k^{(\varepsilon)}\mid \tilde S_1^{(\varepsilon)}=\phi_\varepsilon(s'_1)\big) \rightarrow \log(m_\varepsilon-1), \text{if } p_{\max}^{S}(\varepsilon;s'_1)\to 0,
\end{equation}
which reflects the extreme dispersion dictated by sensitivity.
Therefore, it satisfies:
\begin{equation}
\mathbb{E}_{s'_1 \sim N'(s_1)}\!\big[H_S\big(\tilde S_k^{(\varepsilon)}\mid \tilde S_1^{(\varepsilon)}=\phi_\varepsilon(s_1)\big)\big]
\;-\;
\mathbb{E}_{s'_1 \sim N'(s_1)}\!\big[H_P\big(\tilde S_k^{(\varepsilon)}\mid \tilde S_1^{(\varepsilon)}=\phi_\varepsilon(s_1)\big)\big]
\;\to\; \text{strictly positive}.
\end{equation}

Q.E.D.
\hfill\(\Box\)

\subsection{Consequences for optimal strategy} 
Because the minimum achievable expected NLL equals the conditional entropy, the high conditional entropy of skip-step predictions in serial tasks implies a high irreducible loss for any \(p_\theta(S_k\mid S_1)\) objective. 
A loss-minimizing learner therefore prefers factorizing the prediction into a chain of low-entropy one-step conditionals, i.e., an autoregressive strategy, which aligns with the chain rule factorization and standard maximum-likelihood training. 
By contrast, in parallel tasks, the existence of a stable high-probability outcome for some downstream state \(S_k\) produces a low-entropy, high-confidence conditional, so optimizing \(p_\theta(S_k\mid S_1)\) can be preferable and can support non-autoregressive or partially parallel generation plans.\vspace{-16pt}

\paragraph{Takeaway.}
Task-intrinsic sensitivity versus invariance dictates the shape of the optimal conditional distributions; via the cross-entropy/KL equivalence, this in turn selects the generation procedure that globally minimizes expected loss, with serial tasks forcing autoregression and parallel tasks permitting advantageous skip-step or parallel predictions.

\begin{figure*}[t]
    \centering
    \includegraphics[width=\textwidth]{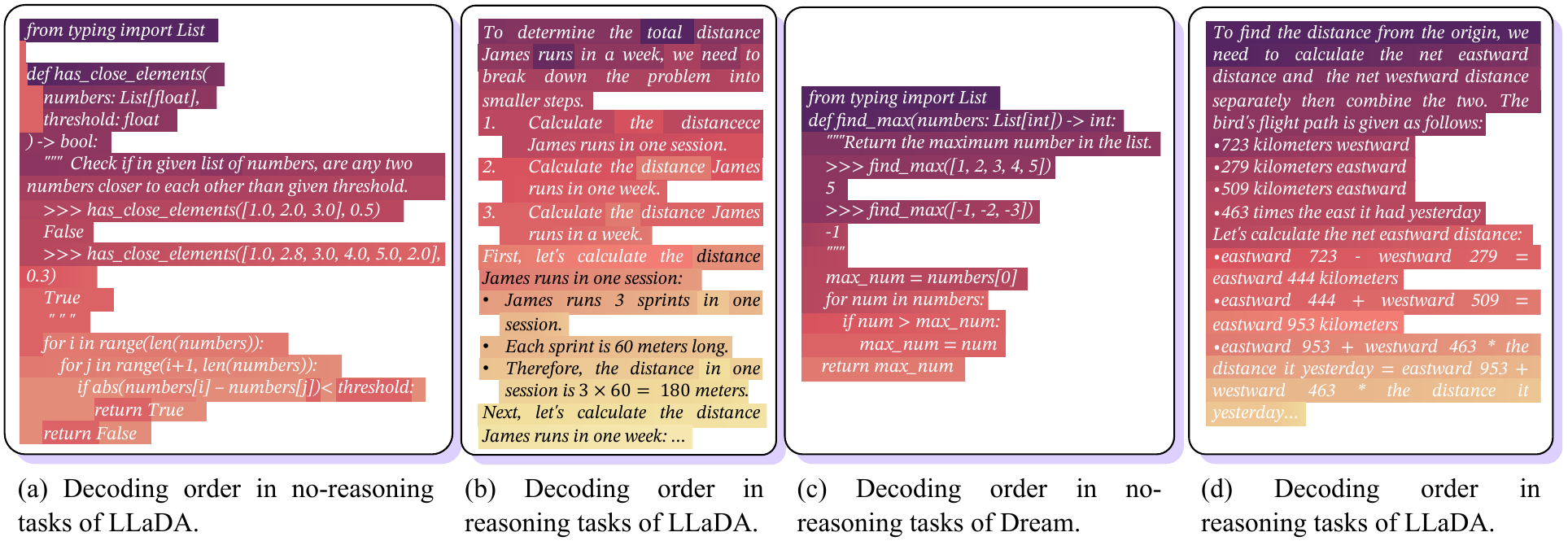}
    \caption{Decoding order of Dream and LLaDA on BigGSM~\citep{chen2024unlocking}.}
    \label{fig:decoding_dream}
\end{figure*}

\section{Diffusion-Step Evaluation Details}
\label{append:diffusion}
In this section, we provide additional technical details on the methodology used to evaluate the impact of diffusion steps and sampling lengths in Diffusion-based Language Models (DLLMs). We specifically focus on how these parameters influence the efficiency and accuracy of the models when tackling complex reasoning tasks.

For our analysis, we utilize the BigGSM dataset~\citep{chen2024unlocking}, which includes a diverse range of complex reasoning tasks designed to test current models' ability to perform long-form reasoning. In particular, we assess the performance of two representative DLLMs on these tasks and compare them against a standard ALLM.
We systematically vary both the number of diffusion steps and the sampling lengths to evaluate their combined effects on the reasoning efficiency of DLLMs. The number of diffusion steps tested ranges from 1 to 1024, while the maximum token lengths vary from 1 to 1024 tokens, with low-confidence remasking. In each experiment, the number of diffusion steps is set equal to the maximum token length. This range allows us to assess the model's performance under different levels of token generation and diffusion refinement. For ALLMs, we adjust the maximum token length between 1 and 1024 and the temperature between 0.2 and 0.7, aiming to achieve comparable performance to that of the DLLMs.

For each setting, we track the following metrics:
\begin{itemize}[leftmargin=16pt, itemsep=0pt, topsep=0pt]  
    \item \textbf{Accuracy:} The percentage of correct answers generated by the model.
    \item \textbf{Model Output Length:} The number of tokens generated by the model before reaching the stopping token (calculated using the GPT-4O tokenizer).
\end{itemize}
When the maximum token length is less than or equal to 512, the model output length typically constitutes 50\% to 80\% of the maximum token length. Specifically, when generating a maximum token length of 512, achieving optimal performance requires 512 diffusion steps combined with low-confidence remasking strategies. We utilize this remasking approach to ensure the best performance.

Our evaluation demonstrates that the efficiency of DLLMs in reasoning tasks is strongly influenced by the balance between diffusion steps and sampling lengths. While a higher number of diffusion steps generally improves reasoning accuracy, it increases computational requirements. Thus, while sufficient diffusion steps are essential for effective reasoning, an excessive number can significantly reduce processing efficiency.

\section{Early-Stop Strategy}

The early stopping mechanism is based on the dynamic stability of tokens, which monitors the variation of 
the newly updated tokens during the diffusion process to judge whether the generation has converged. 
We calculate the overlap ratio between the current step's selected tokens (current\_tokens) and 
the previous step's tokens (prev\_tokens) in each diffusion step. When the overlap ratio of the selected tokens 
remains stable over three consecutive steps and exceeds a threshold $\theta = 0.99$, early stopping is triggered. 
The overlap ratio is calculated as:
\begin{equation}
    \text{overlap\_ratio} = \frac{1}{N} \sum_{j=1}^{N} \mathbb{I}(current\_tokens_j = prev\_tokens_j),
\end{equation}
where $N$ is the number of tokens updated in the current step, and $\mathbb{I}$ is the indicator function. 
This mechanism is controlled by the parameter early\_stop\_threshold = 0.99, which controls 
the sensitivity.The higher the threshold, the more stable the token sequence needs to be before triggering early stopping.

The parameter settings use a block-based diffusion strategy: the total generation length of 512 tokens is divided into blocks of 
length block\_length = 32.
Temperature = 0.7 helps balance exploration and exploitation. We choose low\_confidence strategy, which updates tokens with low confidence. 
This combination ensures the quality of the generated text while improving efficiency 
by using fewer diffusion steps, which typically converge to a value smaller than the maximum 512 steps.

\section{Prompting Experiment Details}
\label{append:prompt}
\subsection{Experimental Setup}
In this study, we employ the following models: Dream-7B-Instruct~\citep{ye2025dream}, LLaDA-8B-Instruct~\citep{nie2025llada}, LLaDA-v1.5~\citep{zhu2025llada}, and LLaDOU-Math~\citep{huang2025reinforcing}. To optimize performance, we experiment with a temperature range of [0, 1], choose top-p=0.95 and block-length=32, and select the maximum token length from the set \{128, 256, 512\}, as well as the diffusion step from \{128, 256, 512\}. For each model, we use the default decoding settings. Additionally, we apply low-confidence remasking to explore the scaling behavior. All experiments conduct on a single A100 or A800 80G GPU.

\subsection{Sequential Reasoning Prompting}
These methods were originally designed primarily for Autoregressive Large Language Models (ALLMs) and have played a  key role in optimizing their reasoning capabilities:

\begin{itemize}[leftmargin=16pt, itemsep=0pt, topsep=0pt]
    \item \textbf{Least-to-Most}~\citep{zhou2022least}: In autoregressive models, this method systematically breaks down complex problems into multiple simpler 
    sub-problems, guiding the model to reason step by step rather than attempting to solve the entire complex problem at once. This effectively 
    reduces the complexity of single-step generation and enhances the model's ability to handle complex reasoning tasks.
    \item \textbf{Zero-CoT}~\citep{NEURIPS2022_8bb0d291}: This strategy uses a simple natural language instruction (e.g., ``Let's think step by step'') to 
    activate the inherent sequential reasoning ability of autoregressive models, enabling them to generate logically coherent reasoning chains without the need for examples. This not only lowers the barrier to prompt design but also significantly improves the zero-shot reasoning performance, 
    enhancing both the efficiency and generality of reasoning.
    \item \textbf{Plan-and-Solve}~\citep{wang2023plan}: By clearly dividing the reasoning process into a planning phase and a detailed execution 
    phase, this strategy helps autoregressive models first construct a solution framework and then fill in specific content. This enhances the structural 
    integrity and global consistency of the solution, proving particularly effective for tasks requiring multi-step logical reasoning and long-range dependency modeling.
\end{itemize}
Together, these strategies strengthen the sequential reasoning ability of autoregressive models, guiding them through prompt design to generate continuous and logically sound reasoning paths in a more systematic and reliable manner.
However, we found that these strategies are not suitable for DLLMs.

\subsection{Constraint-guided Reasoning Prompting}
\paragraph{Complex-CoT:} The original version of Complex-CoT~\citep{fu2022complexity} leverages a few-shot reasoning technique to prompt LLMs into performing more sophisticated reasoning processes. This approach enhances the model’s ability to handle tasks that require a series of logical inferences or multi-step reasoning, thereby improving the overall performance on complex questions. Specifically, by providing a few-shot example that demonstrates how to perform intricate reasoning, the model learns to apply similar patterns to new, unseen problems.

In contrast, the Constrained-Guided Version of Complex-CoT introduces a crucial modification to meet specific requirements. Rather than using few-shot examples, we reframe the prompting method as instruction-based, zero-shot constraints created by human experts. These constraints guide the model’s reasoning process without the need for training on a set of example problems. 
To implement this approach, the following prompting structure is used to ensure that the model approaches each question with the necessary depth and detail:
\begin{figure}[h]
    \centering
    \begin{tcolorbox}[
            colback=blue!5!white,
            colframe=blue!75!black,
            width=\textwidth,
            arc=4pt,
            boxrule=0.5pt,
            title=\textbf{Complex-CoT (Constrained-Guided Version)}
        ]
        \setlength{\parindent}{0pt}
        \setlength{\parskip}{0.8ex}

        You should think about the following question as thoroughly and in as much detail as possible.

        Question: \{question\}
        \end{tcolorbox}\vspace{-16pt}
\end{figure}

\paragraph{MARP:} The original MARP~\citep{chen2024unlocking} employs an instruction-based, in-context-learning approach to guide LLMs in structuring and constraining each step of the reasoning process. This method decomposes complex problems into manageable components by promoting multi-step reasoning, while ensuring each step is focused and achievable. By constraining reasoning at each stage, MARP prevents overgeneralization and ensures logical, organized outputs.

To meet the requirements of the Constrained-Guided Version, we modify MARP in two ways: first, by organizing reasoning into discrete steps, and second, by enabling parallel processing within each step. This approach allows the model to perform multiple operations simultaneously without compromising clarity or precision. The key concept is to balance step-by-step reasoning with parallel processing, enhancing task efficiency. Each reasoning step involves multiple basic operations, ensuring clarity and minimizing computational overhead.

The following prompt structure is used to guide the model’s reasoning process:
\begin{figure}[h]
    \centering
    \begin{tcolorbox}[
            colback=blue!5!white,
            colframe=blue!75!black,
            width=\textwidth,
            arc=4pt,
            boxrule=0.5pt,
            title=\textbf{MARP (Constrained-Guided Version)}
        ]
        \setlength{\parindent}{0pt}
        \setlength{\parskip}{0.8ex}

        Reason step by step, but process operations in parallel.
        \begin{itemize}[leftmargin=16pt, itemsep=0pt, topsep=0pt]
            \item At each step, you may perform multiple simple operations (up to 5).
            \item Each operation must remain basic and not involve excessive complexity.
            \item If you choose to perform more operations in a single step, then each operation must be correspondingly smaller in scope.
        \end{itemize}
        Question: \{question\}
        \end{tcolorbox}\vspace{-16pt}
\end{figure}

\subsection{Parallel-encouraging Prompting}
This parallel-encouraging strategy is essential for reducing PSC in reasoning. To adapt MARP for DLLM, we enable the model to process multiple operations concurrently, avoiding the bottleneck of sequential processing, where each step depends on the completion of the previous one. This parallel processing speeds up reasoning and enhances scalability.
At the same time, operational complexity constraints ensure the reasoning process remains clear and manageable. This method balances parallel execution with simplicity, allowing for effective multi-step reasoning without overwhelming the model with overly complex tasks.
Moreover, the model adjusts operation complexity dynamically. When tasked with more operations in a given step, each operation must be simpler, preventing cognitive overload and helping the model stay focused on individual tasks.

Ultimately, this approach enables the model to execute parallel reasoning efficiently while maintaining clarity and precision. The detailed prompting for implementation is as follows:
\begin{figure}[h]
    \centering
    \begin{tcolorbox}[
            colback=blue!5!white,
            colframe=blue!75!black,
            width=\textwidth,
            arc=4pt,
            boxrule=0.5pt,
            title=\textbf{Diff-MARP}
        ]
        \setlength{\parindent}{0pt}
        \setlength{\parskip}{0.8ex}

        Reasoning in parallel. In each step, do as many basic operations as you can, up to 5.
        
        Any single operation cannot be too complex.
        
        If you use more operations in a step, the maximum allowed size for any operation decreases.
        
        Question: \{question\}
        \end{tcolorbox}\vspace{-16pt}
\end{figure}

\section{DLLM's Limited Capabilities of Long CoT Reasoning}
\label{append:capability}
\subsection{Long Chain-of-Thought Capabilities}
Following \citet{chen2025towards}, the Long Chain-of-Thought (Long CoT) reasoning capabilities comprise three linked components: deep reasoning, exploration, and reflection.\vspace{-16pt}

\paragraph{Deep Reasoning.}  Given $s_i$ as the $i$-th reasoning step, Deep reasoning models the conditional probability $p_\theta(s_0, s_1, \dots, s_K|s_0)$, facilitating multi-step logical inference through iterative refinement. The associated reverse process can be characterized by a factorization:
\begin{equation}
p_\theta(s_0, s_1, \dots, s_K|s_0) = \prod_{i=0}^K p_\theta(s_{i+1} | s_i).\vspace{-16pt}
\end{equation}

\paragraph{Exploration.} Exploration stems from the probabilistic nature of the reverse process. At each exploration step $s_j$, multiple samples $s_{j}^k$ can be drawn from the conditional distribution $p_\theta(s_{j} | s_i, i < j)$, enabling the model to explore diverse plausible continuations or solutions. This is formalized as:
\begin{equation}
s_{j}^k \sim p_\theta(s_{j} | s_i, i < j), \quad k = 1, \ldots, K,
\end{equation}
where $K$ controls the breadth of exploration. This sampling diversity enhances robustness by covering multiple reasoning paths and mitigating premature convergence to suboptimal outputs.
\vspace{-16pt}

\paragraph{Reflection.} We view reflection as a self-correction mechanism arising from iterative conditioning on latent states. At each reverse step, the model revises its belief about the target sequence using the immediately previous state and, via the accumulated latent trajectory, all prior estimates. Formally, this corresponds to implicit message passing:
\begin{equation}
    \hat{s}_{j} \sim p_\theta(s_{j} | s_i, i \ge j), 
\end{equation}
where $\hat{s}_{j}$ denotes the corrected state at step $j$, enabling iterative error correction and refinement.

Together, these components yield a procedure that combines structured logics with stochastic exploration and continual self-correction, enabling effective reasoning on complex multi-step tasks.

\subsection{Stategies for Self-Reflection and Self-Exploration experiments on DLLMs} 

To investigate whether DLLMs truly possess the fundamental capabilities for Long CoT Reasoning, we designed two sets of experiments: self-reflection 
and self-exploration,using two distinct prompting strategies to examine the basic abilities of DLLMs.\vspace{-16pt}

\paragraph{Self-Reflection:}
(1)Prompting Reflection, structured reflection prompts are embedded within initial instructions,requiring the model to 
perform logical self-checking during generation. 
(2)Autoregressive Forcing Reflection, correction prompts (e.g., "Wait...there might be something wrong") are replaced with the
end-of-sequence (EOS) token as a post-generation intervention strategy.\vspace{-16pt}

\paragraph{Self-Exploration:} (1) Prompting Exploration, which embeds exploration prompts in initial instructions to activate multi-path reasoning.
(2) Autoregressive Forcing Exploration, which replaces EOS token to “Let’s think in another way...” to induce exploratory reasoning.

\subsection{Evaluation of Self-Reflection and Self-Exploration Capabilities} 

In evaluating the self-reflection capabilities of the LLaDA-8B-Instruct~\citep{nie2025large} and Dream-7B-Instruct~\citep{ye2025dream} models, the BigGSM dataset was utilized. During the generation process, we employed a temperature of 0.7 for self-reflection and 0.2 for self-exploration, coupled with top-p sampling set to 0.95. Additionally, diffusion steps were configured to 512, and the generation length was fixed at 512.
For the investigation into the models' self-exploration capabilities, the experimental settings were identical, with the sole distinction being the substitution of the reflection strategy with an exploration strategy. Based on the setting of \citet{qin2023cross}, we utilize the following reasoning metrics for deeper analysis:\vspace{-16pt}

\paragraph{Semantic Alignment:}
The semantic alignment metrics~\citep{golovneva2022roscoe} lies in the \textit{reasoning alignment vector}, which spans from the $N$-step hypothesis $\boldsymbol{h}$ to the source $\boldsymbol{s}$ of length $T$:
\begin{equation}
r\text{-}align(\boldsymbol{h}\to\boldsymbol{s})=\{\alpha_{1},\alpha_{2},\cdots,\alpha_{N}\},
\end{equation}
where each alignment value can be calculated as:
\begin{equation}
\alpha_{i}=r\text{-}align(h_{i}\to\boldsymbol{s})=\frac{\left[1+\max_{j=1}^{T}\cos(h_{i},s_{j})\right]}{2} \in [0,1].
\end{equation}
Here, such an alignment value is the normalized cosine similarity between the reference step and the most similar sentence in a context, and explicitly measures the \textit{grounding} of the step-wise reasoning with respect to the source text.
The alignment vector \( r\text{-align}(h \rightarrow s) \) is estimated by matching the source text and the reasoning chain on the embeddings of the tokens and individual reasoning steps. A similar confidence alignment score is introduced in CTC to measure whether the information of the \( i \)-th source document token \( s_j \) is supported by the hypothesis token \( h_i \), assessing whether the reasoning step \( h_i \) supports the source context \( s \).\vspace{-16pt}

\paragraph{Repetition-word:}
To identify repeated, or paraphrased steps, we look at the repetition word scores~\citep{golovneva2022roscoe} between all steps in the hypothesis chain: 
\[
1-\max_{i=2\ldots N}\max_{j=1\ldots i-1}\left[\left(1/M_{i}\right)\sum_{l=1}^{M_{i}}r_{\text{align}}^{\text{token}}(h_{i,l}\to h_{j})\right].
\]
For each pair of sentences, we look at the mean token alignment and find those sentences that maximize this alignment score. In other words, Repetition-Token will punish chains where there are at least two steps with high overlap in token embeddings.\vspace{-16pt}

\paragraph{Informativeness:} 
Measures how well information present in the source is used in the reasoning steps, we calculate informativeness~\citep{golovneva2022roscoe}: 
\[
\frac{(1/T)\sum_{t=1}^{T}r_{\text{align}}(s_{t}\to h) + (1/N)\sum_{i=1}^{N}r_{\text{align}}(h_{i}\to s)}{2}.
\]
Info-step gives a higher score to reasoning steps that are well-grounded with respect to the source, and identifies the degree of information from source that is covered by the generated hypothesis. A lower Info-Step score corresponds to the reasoning steps that are not related to the source sentences or have missed information provided in the context.\vspace{-16pt}

\paragraph{Reasoning-Alignment}:  
The most straightforward way to evaluate the correctness of the hypothesis chain is to compare the degree of the overlap between the hypothesis and the reference. One way of doing that is to measure the reasoning alignment~\citep{golovneva2022roscoe} between them:
\[
\frac{1}{N} \sum_{i=1}^{N} r_{\text{align}}(h_{i} \to r).
\]\vspace{-16pt}

\paragraph{Token-Entropy}:  
To calculate the token entropy, we will utilize the pipeline as follows:
First, calculate the probability of each token \( p(t_i) \), which is the frequency of token \( t_i \) divided by the total number of tokens \( N \):
\[
p(t_i) = \frac{\text{count}(t_i)}{N}
\]
Next, calculate the information content \( I(t_i) \) of each token, which reflects the uncertainty contribution of that token to the text:
\[
I(t_i) = -\log(p(t_i))
\]
Finally, token-entropy is the weighted average of the information content of all tokens, given by:
\[
H = - \sum_{i=1}^{N} p(t_i) \log(p(t_i))
\]
where \( p(t_i) \) is the probability of token \( t_i \), and \( \log(p(t_i)) \) is the corresponding logarithmic information content.
Token-entropy reflects the overall uncertainty of the text. A higher value indicates that the text is more random and diverse, while a lower value suggests that the text is more focused and repetitive.\vspace{-16pt}

\paragraph{Cosine similarity (Sim)} 
Cosine similarity measures the degree of similarity between two vectors encoded by BGE~\citep{bge_embedding} by calculating the cosine of the angle between them. For text embedding vectors, a value closer to 1 indicates greater semantic similarity.
Let the two generated text vectors be \( \mathbf{A} \) and \( \mathbf{B} \). The formula for calculating their cosine similarity is:
\begin{equation}
    \text{cosine\_similarity}(\mathbf{A}, \mathbf{B}) = \frac{\mathbf{A} \cdot \mathbf{B}}{\| \mathbf{A} \| \cdot \| \mathbf{B} \|} = \frac{\sum_{i=1}^{n} A_i B_i}{\sqrt{\sum_{i=1}^{n} A_i^2} \cdot \sqrt{\sum_{i=1}^{n} B_i^2}}
\end{equation}
where \( \mathbf{A \cdot B} \) is the dot product of vectors \( \mathbf{A} \) and \( \mathbf{B} \). \( \| \mathbf{A} \| \) and \( \| \mathbf{B} \| \) are the Euclidean norms (magnitudes) of vectors \( \mathbf{A} \) and \( \mathbf{B} \).
\( A_i \) and \( B_i \) represent the components of vectors \( \mathbf{A} \) and \( \mathbf{B} \) along the \( i \)-th dimension.\vspace{-16pt}

\paragraph{Perplexity (PPL) of the Model} 

Perplexity is a concept in information theory used to measure the uncertainty of a probabilistic model in predicting samples. In natural language processing, it is employed to evaluate how well a language model fits a set of test data.

Given a sequence of \( N \) tokens \( W = w_1, w_2, \dots, w_N \), where the language model predicts the probability of this sequence \( P(W) \), the perplexity of the sequence is defined as:
\begin{equation}
\text{PPL}(W) = P(W)^{-\frac{1}{N}} = \exp \left( -\frac{1}{N} \log P(W) \right).
\end{equation}
Because of the sequence's independence assumption, we can compute \( P(W) \) as:
\begin{equation}
P(W) = \prod_{i=1}^{N} P(w_i | w_1, \dots, w_{i-1}).
\end{equation}
Therefore, the commonly seen formula for perplexity is:
\begin{equation}
\text{PPL}(W) = \exp \left( -\frac{1}{N} \sum_{i=1}^{N} \log P(w_i | w_1, \dots, w_{i-1}) \right),
\end{equation}
where \( \log P(W) \) is the log probability of the entire sequence. \( \frac{1}{N} \sum_{i=1}^{N} \log P(w_i | w_1, \dots, w_{i-1}) \) is the average log probability of the sequence under the model. \( \exp \) is the exponential function, used to transform the log probability back to its original scale.
        
Observation of the results for both DREAM~\citep{ye2025dream} and LLaDA~\citep{nie2025large} models, under both self-reflection and self-exploration settings, the scores across all ROSCOE-SA evaluation metrics are highly similar. This indicates that current diffusion language models (DLLMs) have not yet genuinely acquired the deeper capabilities of self-reflection and self-exploration, as their outputs do not exhibit significant differences under varying strategic prompts.

\subsection{Evaluation of Deep-Reasoning Capability}
Following \citet{chen2024unlocking}, we further investigate reasoning boundaries (RBs) in deep reasoning capabilities in mathematical reasoning. We prompt DLLMs to generate plans and assess their accuracy through manual evaluation. When the model meets the question with fewer than 1 reasoning steps, accuracy surpasses 80\%. Conversely, when reasoning steps exceed 3-4, accuracy falls below 10\%. Moreover, we first randomly select 200 samples to generate examples and split steps from the DLLM-generated rationales based on ROSCOE~\citep{golovneva2022roscoe}. Further, we also manually identify the first model's incorrect step position.

\section{Three-directional Inference-Time Scaling on DLLMs}
\label{append:scaling}

\begin{figure*}[t]
    \centering
    \includegraphics[width=\textwidth]{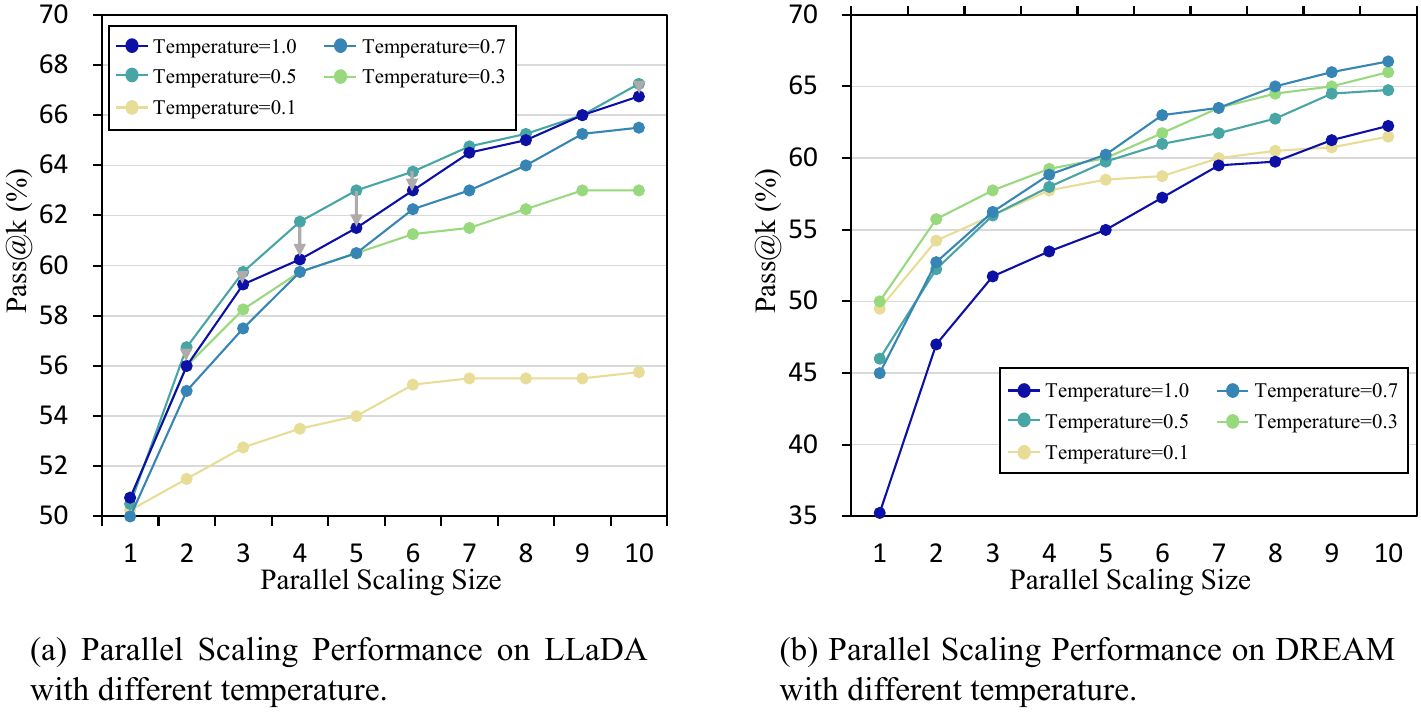}
    \caption{Parallel scaling performance of DLLMs under Different Temperature Settings}
    \label{fig:appendix_parallel-scaling}
\end{figure*}
\subsection{Parallel Scaling Experiment Details}
In the parallel scaling section, we utilize the dual-cache generation strategy from Fast-dLLM based on the diffusion language model 
LLaDA-8B-Instruct~\citep{nie2025large}, and perform batch processing on the BigGSM~\citep{chen2024unlocking} dataset. Key configurations include:diffusion\_steps=256, gen\_length=256, block\_length=32, and a Dynamic Low-Confidence Remasking mechanism.

We also employed the dual\_cache generation strategy from Fast-dLLM~\citep{wu2025fast} on the Dream-7B-Instruct~\citep{ye2025dream} model for testing on the  BigGSM reasoning dataset. 
The core configuration includes: diffusion\_steps=256, gen\_length=256, and block\_length=32.

The results are shown in Figure~\ref{fig:appendix_parallel-scaling}.
It can be observed that the accuracy generally increases with higher k-values. At the initial attempts, the accuracy at Temperature 1.0 
was relatively low. Although it showed significant improvement in the early stages, its later accuracy fell behind other temperatures. 
At Temperature 0.1, the accuracy growth was more stable initially, but eventually plateaued at around 60\%, similar to Temperature 1.0. Overall, intermediate temperatures demonstrated better pass@k accuracy performance, achieving higher accuracy with more consistent and stable growth.

\subsection{Diffusion Scaling Experiment Details}
We set the diffusion\_step to be between 1 and 4096 (with max-token-length equal to 512). The other settings are identical to those of parallel scaling. The results are shown in Figure~\ref{fig:diffusion-scaling} (a).

\subsection{Sequential Scaling Experiment Details}
We set the Max Token Length to be between 1 and 4096 (with diffusion-step equal to max-token-length). The other settings are identical to those of parallel scaling. The results are shown in Figure~\ref{fig:diffusion-scaling} (d).

\end{document}